\documentclass[runningheads]{llncs}
\usepackage{graphicx}

\usepackage{tikz}
\usepackage{comment}
\usepackage{amsmath,amssymb} 
\usepackage{color}
\usepackage{cite}
\usepackage[ruled,vlined,noend]{algorithm2e}
\usepackage{algpseudocode}
\usepackage{hyperref}
\usepackage{makecell}

\DeclareMathOperator*{\argmax}{arg\,max}
\DeclareMathOperator*{\argmin}{arg\,min}
\DeclareMathOperator*{\bce}{BCE}
\DeclareMathOperator*{\iou}{IoU}
\DeclareMathOperator*{\nms}{NMS}
\DeclareMathOperator*{\oks}{OKS}

\newcommand{\etal}{\textit{et al.}}
\newcommand{\eg}{\textit{e.g.}}
\newcommand{\ie}{\textit{i.e.}}
\newcommand{\etc}{\textit{etc.}}

\newcommand\hll[1]{%
  \bgroup
  \hskip0pt\color{black}%
  #1%
  \egroup
}
\newcommand\hlll[1]{%
  \bgroup
  \hskip0pt\color{black}%
  #1%
  \egroup
}

\usepackage[accsupp]{axessibility}  

\begin{document}
\pagestyle{headings}
\mainmatter
\def\ECCVSubNumber{4439}  

\title{Rethinking Keypoint Representations: Modeling Keypoints and Poses as Objects for Multi-Person Human Pose Estimation} 

\titlerunning{KAPAO: Keypoints and Poses as Objects}
%
\author{William McNally\orcidID{0000-0002-7187-7147} \and
Kanav Vats\orcidID{0000-0002-9515-7259} \and
Alexander Wong\orcidID{0000-0002-5295-2797} \and
John McPhee\orcidID{0000-0003-3908-9519}} 
\authorrunning{W. McNally et al.}
%
\institute{Systems Design Engineering, University of Waterloo, Canada\\
Waterloo Artificial Intelligence Institute, University of Waterloo, Canada\\
\email{\{wmcnally, k2vats, a28wong, mcphee\}@uwaterloo.ca}
}
\maketitle

\begin{abstract}
In keypoint estimation tasks such as human pose estimation, heatmap-based regression is the dominant approach despite possessing notable drawbacks: heatmaps intrinsically suffer from quantization error and require excessive computation to generate and post-process. Motivated to find a more efficient solution, we propose to model individual keypoints and sets of spatially related keypoints (\ie, poses) as objects within a dense single-stage anchor-based detection framework. Hence, we call our method KAPAO (pronounced ``Ka-Pow''), for \textbf{K}eypoints \textbf{A}nd \textbf{P}oses \textbf{A}s \textbf{O}bjects. KAPAO is applied to the problem of single-stage multi-person human pose estimation by simultaneously detecting human pose and keypoint objects and fusing the detections to exploit the strengths of both object representations. In experiments we observe that KAPAO is faster and more accurate than previous methods, which suffer greatly from heatmap post-processing. The accuracy-speed trade-off is especially favourable in the practical setting when not using test-time augmentation. Source code: \url{https://github.com/wmcnally/kapao}.
\end{abstract}

\section{Introduction}
Keypoint estimation is a computer vision task that involves localizing points of interest in images. It has emerged as one of the most highly researched topics in the computer vision literature~\cite{mcnally2021evopose2d, pavllo20193d, iqbal2018hand, huang2020awr, mcnally2018action, mcnally2019star, gavrilyuk2020actor, zhou2019objects, andriluka2018posetrack, raaj2019efficient, dong2018style, wang2019adaptive, xu2021anchorface, suwajanakorn2018discovery, jakab2018unsupervised, voeikov2020ttnet, mcnally2021deepdarts}. The most common method for estimating keypoint locations involves generating target fields, referred to as \textit{heatmaps}, \hlll{that center} 2D Gaussians on the target keypoint coordinates. Deep convolutional neural networks~\cite{lecun1995convolutional} are then used to regress the target heatmaps on the input images, and keypoint predictions are made via the arguments of the maxima of the predicted heatmaps~\cite{tompson2014joint}. 

While strong empirical results have positioned heatmap regression as the \textit{de facto} standard method for detecting and localizing keypoints~\cite{tompson2014joint, newell2016stacked, cao2017realtime, chen2018cascaded, xiao2018simple, sun2019deep, mcnally2021evopose2d, yang2021transpose, cheng2020higherhrnet, geng2021bottom, khirodkar2021multi, luo2021rethinking, braso2021center}, there are several known drawbacks. First, these methods suffer from quantization error\hlll{:} the precision of a keypoint prediction is inherently limited by the spatial resolution of the \hlll{output} heatmap. Larger heatmaps are therefore advantageous, but require additional upsampling operations and costly processing at higher resolution~\cite{mcnally2021evopose2d, cheng2020higherhrnet, geng2021bottom, luo2021rethinking, braso2021center}. Even when large heatmaps are used, special post-processing steps are required to refine keypoint predictions, \hlll{slowing} down inference~\cite{newell2016stacked, chen2018cascaded, cheng2020higherhrnet, luo2021rethinking}. Second, when two keypoints of the same type (\ie, class) appear in close proximity to one another, the overlapping heatmap signals may be mistaken for a single keypoint. Indeed, this is a common failure case~\cite{cao2017realtime}. For these reasons, researchers have started \hlll{to investigate} alternative, \textit{heatmap-free} keypoint detection methods~\cite{mcnally2021deepdarts, li20212d, li2021pose, li2021human, xu2021anchorface}.

\begin{figure}[t!]
\centering
    \includegraphics[trim={0.2cm, 0cm, 0cm, 0cm}, clip, width=0.6\linewidth]{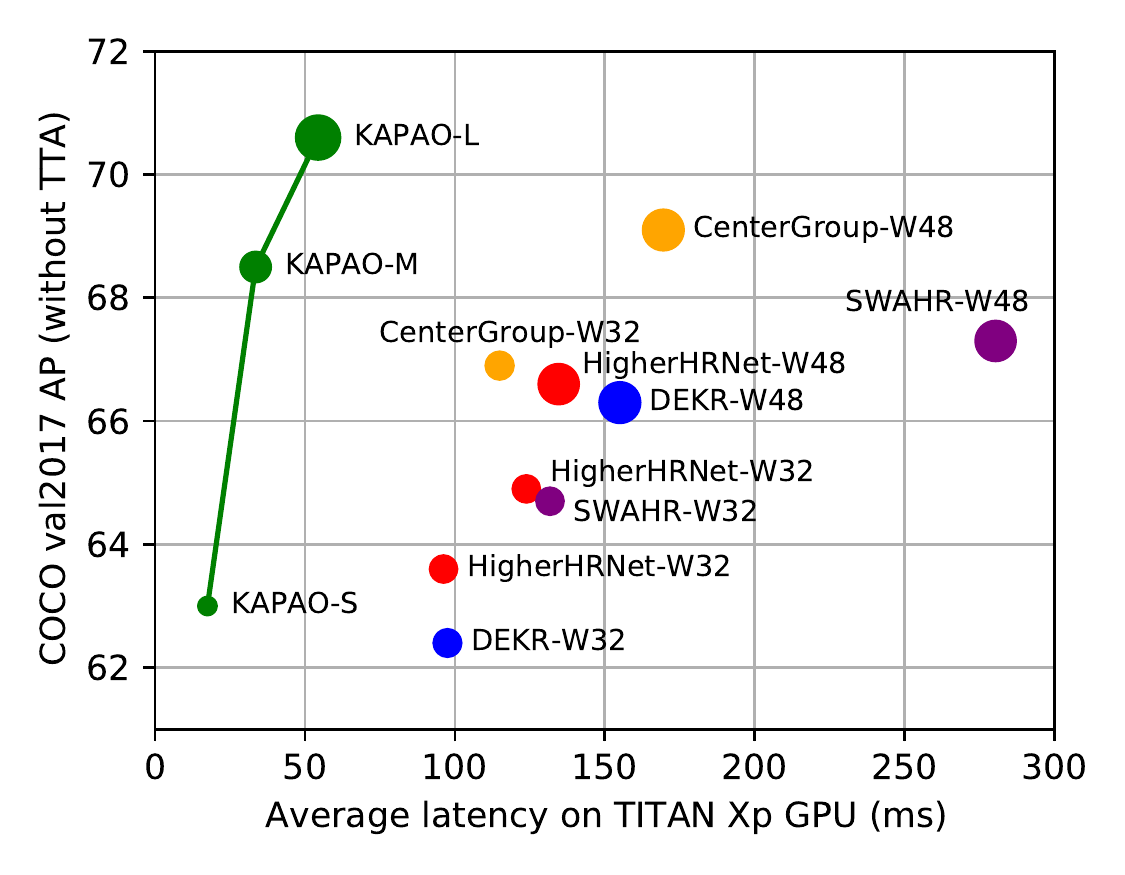}
\caption[Accuracy vs. Inference Speed: KAPAO compared to state-of-the-art single-stage methods.]{Accuracy vs. Inference Speed: KAPAO compared to state-of-the-art single-stage multi-person human pose estimation methods DEKR~\cite{geng2021bottom}, HigherHRNet~\cite{cheng2020higherhrnet}, HigherHRNet + SWAHR~\cite{luo2021rethinking}, and CenterGroup~\cite{braso2021center} without test-time augmentation (TTA), \ie, excluding multi-scale testing and horizontal flipping. The raw data are provided in Table~\ref{tab:kapao_coco_val}. The circle size is proportional to the number of model parameters.}
\label{fig:acc_speed}
\end{figure}

In this paper, we introduce a new heatmap-free keypoint detection method and apply it to single-stage multi-person human pose estimation. Our method builds on recent research showing how keypoints can be modeled as objects within a dense anchor-based detection framework by representing keypoints at the center of small \textit{keypoint bounding boxes}~\cite{mcnally2021deepdarts}. In preliminary experimentation with human pose estimation, we found that this keypoint detection approach works well for human keypoints that are characterized by local image features (\eg, the eyes), but the same approach is less effective at detecting human keypoints that require a more global understanding (\eg, the hips). We therefore introduce a new \textit{pose object} representation to help detect sets of keypoints that are spatially related. Furthermore, we detect keypoint objects and pose objects simultaneously and fuse the results using a simple matching algorithm to exploit the benefits of both object representations. By virtue of detecting pose objects, we unify person detection and keypoint estimation \hlll{and provide} a highly efficient single-stage approach to multi-person human pose estimation.

As a result of not using heatmaps, KAPAO compares favourably against recent single-stage human pose estimation models in terms of accuracy \textit{and} inference speed, especially when not using test-time augmentation (TTA), which \hlll{represents} how such models are deployed in practice. As shown in Figure \ref{fig:acc_speed}, KAPAO achieves an AP of 70.6 on the Microsoft COCO Keypoints validation set without TTA \hlll{while having} an average latency of 54.4 ms (forward pass + post-processing time). Compared to the state-of-the-art single-stage model HigherHRNet + SWAHR~\cite{luo2021rethinking}, KAPAO is 5.1$\times$ faster and 3.3 AP more accurate when not using TTA. Compared to CenterGroup~\cite{braso2021center}, KAPAO is 3.1$\times$ faster and 1.5 AP more accurate. 
The contributions of this work are summarized \hlll{as follows}:
\begin{itemize}
    \item A new \textit{pose object} representation is proposed that extends the conventional object representation by including a set of keypoints associated with the object. 
    \smallskip
    \item A new approach to single-stage human pose estimation is developed by simultaneously detecting \textit{keypoint objects} and \textit{pose objects} and fusing the detections. The proposed heatmap-free method is significantly faster and more accurate than state-of-the-art heatmap-based methods when not \hlll{using} TTA.
\end{itemize}

\section{Related Work}
\smallskip\noindent\textbf{Heatmap-free keypoint detection.}
DeepPose~\cite{toshev2014deeppose} regressed keypoint coordinates directly from images using a cascade of deep neural networks that iteratively refined the keypoint predictions. Shortly thereafter, Tompson \etal~\cite{tompson2014joint} introduced the notion of keypoint heatmaps, which have since remained prevalent in human pose estimation~\cite{wei2016convolutional, newell2016stacked, cao2017realtime, chen2018cascaded, xiao2018simple, sun2019deep, cheng2020higherhrnet, geng2021bottom, mcnally2021evopose2d, khirodkar2021multi, yang2021transpose} and other keypoint \hlll{detection} applications~\cite{iqbal2018hand, dong2018style, wang2019adaptive, huang2020awr, vats2019pucknet}. Remarking the computational inefficiencies associated with generating heatmaps, Li \etal~\cite{li20212d} disentangled the horizontal and vertical keypoint coordinates such that each coordinate was represented using a one-hot encoded vector. \hlll{This} saved computation and permitted an expansion of the output resolution, thereby reducing the effects of quantization error and eliminating the need for refinement post-processing. Li \etal~\cite{li2021human} introduced the residual log-likelihood (RLE), a novel loss function for direct keypoint regression based on normalizing flows~\cite{rezende2015variational}. Direct keypoint regression has also been attempted using Transformers~\cite{li2021pose}. 


Outside the realm of human pose estimation, Xu \etal~\cite{xu2021anchorface} regressed anchor templates of facial keypoints and aggregated them to achieve state-of-the-art accuracy in facial alignment. In sports analytics, McNally \etal~\cite{mcnally2021deepdarts} encountered the issue of overlapping heatmap signals in the development of an automatic scoring system for darts and therefore opted to model keypoints as objects using small square bounding boxes. This keypoint representation proved to be highly effective and serves as the inspiration for \hlll{this} work. 

\smallskip\noindent\textbf{Single-stage human pose estimation.}
Single-stage human pose estimation methods predict the poses of every person in an image using a single forward pass~\cite{cao2017realtime, newell2017associative, he2017mask, kreiss2019pifpaf, cheng2020higherhrnet,geng2021bottom, nie2019single}. In contrast, two-stage methods~\cite{papandreou2017towards, fang2017rmpe, chen2018cascaded, xiao2018simple, sun2019deep, mcnally2021evopose2d, khirodkar2021multi, li2021human} first detect \hlll{the} people in an image using an off-the-shelf person detector (\eg, Faster R-CNN~\cite{ren2015faster}, YOLOv3~\cite{redmon2018yolov3}, etc.) and then estimate poses for each detection. Single-stage methods are generally less accurate, but usually perform better in crowded scenes~\cite{li2019crowdpose} and are often preferred because of their simplicity and efficiency, which becomes particularly favourable as the number of people in the image increases. Single-stage approaches vary more in their design compared to two-stage approaches. For instance, they may: (i) detect all the keypoints in an image and perform a \textit{bottom-up} grouping into human poses~\cite{pishchulin2016deepcut, insafutdinov2016deepercut, iqbal2016multi, cao2017realtime, newell2017associative, cheng2020higherhrnet, kreiss2019pifpaf, jin2020differentiable, braso2021center, luo2021rethinking}; (ii) extend object detectors to unify person detection and keypoint estimation~\cite{he2017mask, wei2020point, zhou2019objects, mao2021fcpose}; or (iii) use alternative keypoint/pose representations (\eg, predicting root keypoints and relative displacements~\cite{papandreou2018personlab, nie2019single, geng2021bottom}). We briefly summarize the most recent state-of-the-art single-stage methods below. 

Cheng \etal~\cite{cheng2020higherhrnet} repurposed HRNet~\cite{sun2019deep} for bottom-up human pose estimation by adding a transpose convolution to double the output heatmap resolution (HigherHRNet) and using associative embeddings~\cite{newell2017associative} for keypoint grouping. They also implemented multi-resolution training to address the scale variation problem.
Geng~\etal~\cite{geng2021bottom} predicted person center heatmaps and $2K$ offset maps representing offset vectors for the $K$ keypoints of a pose candidate centered on each pixel using an HRNet backbone. They also disentangled the keypoint regression (DEKR) using separate regression heads and adaptive convolutions.
Luo~\etal~\cite{luo2021rethinking} used HigherHRNet as a base and proposed scale and weight adaptive heatmap regression (SWAHR), which scaled the ground-truth heatmap Gaussian \hlll{variances} based on the person scale and balanced the foreground/background loss weighting. Their modifications provided significant accuracy improvements over HigherHRNet and comparable performance to many two-stage methods.
Again using HigherHRNet as a base, Bras\'{o}~\etal~\cite{braso2021center} proposed CenterGroup to match keypoints to person centers using a fully differentiable self-attention module that was trained end-to-end together with the keypoint detector. \hlll{Notably, all of the aforementioned methods} suffer from costly heatmap post-processing and as such, their inference speeds leave much to be desired.

\smallskip\noindent\textbf{Extending object detectors for human pose estimation.} There \hlll{is} significant overlap between the tasks of object detection and human pose estimation. For instance, He~\etal~\cite{he2017mask} used the Mask R-CNN instance segmentation model for human pose estimation by predicting keypoints using one-hot masks. Wei~\etal~\cite{wei2020point} proposed Point-Set Anchors, which adapted the RetinaNet~\cite{lin2017focal} object detector using pose anchors instead of bounding box anchors. Zhou \etal~\cite{zhou2019objects} modeled objects using heatmap-based center points with CenterNet and represented poses as a 2K-dimensional property of the center point. Mao~\etal~\cite{mao2021fcpose} adapted the FCOS~\cite{tian2019fcos} object detector with FCPose using dynamic filters~\cite{jia2016dynamic}. While these methods based on object detectors provide good efficiency, their accuracies have not competed with state-of-the-art heatmap-based methods. Our work is most similar \hlll{to} Point-Set Anchors~\cite{wei2020point}; however, our method does not require defining data-dependent pose anchors. Moreover, we simultaneously detect individual keypoints and poses and fuse the detections to improve the accuracy of our final pose predictions.

\section{KAPAO: Keypoints and Poses as Objects}
KAPAO uses a dense detection network to simultaneously predict a set of keypoint objects $\{\hat{\mathcal{O}}^k\in\hat{\mathbf{O}}^k\}$ and a set of pose objects $\{\hat{\mathcal{O}}^p\in\hat{\mathbf{O}}^p\}$, collectively $\hat{\mathbf{O}} = \hat{\mathbf{O}}^k\bigcup\hat{\mathbf{O}}^p$. We introduce the concept behind each object type \hlll{and} the relevant notation below. All units are assumed to be in pixels unless stated otherwise.

A keypoint object $\mathcal{O}^k$ is an adaptation of the conventional object representation in which the coordinates of a keypoint are represented at the center $(b_x, b_y)$ of a small bounding box $\mathbf{b}$ with equal width $b_w$ and height $b_h$: $\mathbf{b} = (b_x, b_y, b_w, b_h)$. The hyperparameter $b_s$ controls the keypoint bounding box size (\ie, $b_s$ = $b_w$ = $b_h$). There are $K$ classes of keypoint objects, one for each type in the dataset~\cite{mcnally2021deepdarts}. 

Generally speaking, a pose object $\mathcal{O}^p$ is considered to be an extension of the conventional object representation that additionally includes a set of keypoints associated with the object. While we expect pose objects to be useful in related tasks such as facial and object landmark detection~\cite{xu2021anchorface,jeon2019joint}, they are applied herein to human pose estimation via detection of \textit{human pose objects}, comprising a bounding box of class ``person,'' and a set of keypoints $\mathbf{z} = \{(x_k, y_k)\}_{k=1}^K$ that coincide with anatomical landmarks.

Both object representations possess unique advantages. Keypoint objects are specialized for the detection of individual keypoints that are characterized by strong local features. Examples of such keypoints that are common in human pose estimation include the eyes, ears and nose. However, keypoint objects carry no information regarding the concept of a person or pose. If used on their own for multi-person human pose estimation, a bottom-up grouping method would be required to parse the detected keypoints into human poses. In contrast, pose objects are better suited for localizing keypoints with weak local features as they enable the network to learn the spatial relationships within a set of keypoints. Moreover, they can be leveraged for multi-person human pose estimation directly without the need for bottom-up keypoint grouping.

Recognizing that keypoint objects exist in a subspace of a pose objects, the KAPAO network was designed to simultaneously detect both object types with minimal computational overhead using a single shared network head. During inference, the more precise keypoint object detections are fused with the human pose detections using a simple tolerance-based matching algorithm that improves the accuracy of the human pose predictions without sacrificing any significant amount of inference speed. The following sections provide details on the network architecture, the loss function used to train the network, and inference.

\subsection{Architectural Details}
\label{sec:kapao_arch}
\begin{figure}[t]
\centering
    \includegraphics[trim={0cm, 0cm, 0cm, 0cm}, clip, width=\linewidth]{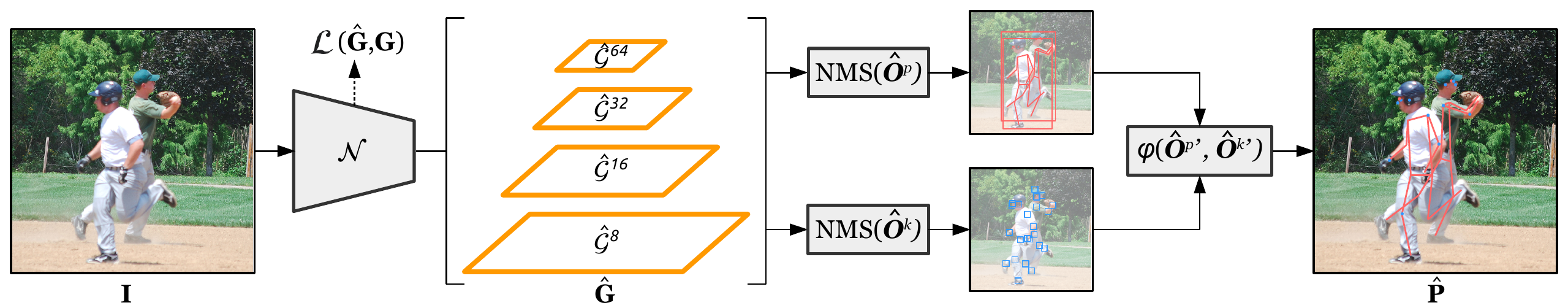}
\caption[Schematic of the KAPAO methodology.]{KAPAO uses a dense detection network $\mathcal{N}$ trained using the multi-task loss $\mathcal{L}$ to map an RGB image $\mathbf{I}$ to a set of output grids $\hat{\mathbf{G}}$ containing the predicted pose objects $\hat{\mathbf{O}}^p$ and keypoint objects $\hat{\mathbf{O}}^k$. Non-maximum suppression (NMS) is used to obtain candidate detections $\hat{\mathbf{O}}^{p\prime}$ and $\hat{\mathbf{O}}^{k\prime}$, which are fused together using a matching algorithm $\varphi$ to obtain the final human pose predictions $\hat{\mathbf{P}}$. The $N_a$ and $N_o$ dimensions in $\hat{\mathbf{G}}$ are not shown for clarity.}
\label{fig:kapao_method}
\end{figure}

A diagram of the KAPAO pipeline is provided in Figure~\ref{fig:kapao_method}. It uses a deep convolutional neural network $\mathcal{N}$ to map an RGB input image $\mathbf{I}\in\mathbb{R}^{h\times w\times 3}$ to a set of four output grids $\hat{\mathbf{G}} = \{\hat{\mathcal{G}}^s\mid s\in\{8, 16, 32, 64\}\}$ containing the object predictions $\hat{\mathbf{O}}$, where $\hat{\mathcal{G}}^s\in\mathbb{R}^{\frac{h}{s}\times \frac{w}{s}\times N_a \times N_o}$:
\begin{equation}
    \mathcal{N}(\mathbf{I}) = \hat{\mathbf{G}}.
\end{equation}
$N_a$ is the number of anchor channels and $N_o$ is the number of output channels for each object. $\mathcal{N}$ is a YOLO-style feature extractor that makes extensive use of Cross-Stage-Partial (CSP) bottlenecks~\cite{wang2020cspnet} within a feature pyramid~\cite{lin2017feature} macroarchitecture. To provide flexibility for different speed requirements, three sizes of KAPAO models were trained (\ie, KAPAO-S/M/L) by scaling the number of layers and channels in $\mathcal{N}$. 

Due to the nature of strided convolutions, the features in an output grid cell $\hat{\mathcal{G}}^s_{i,j}$ are conditioned on the image patch $\mathbf{I}_p=\mathbf{I}_{si:s(i+1), sj:s(j+1)}$. Therefore, if the center of a target object $(b_x, {b_y})$ is situated in $\mathbf{I}_p$, the output grid cell $\hat{\mathcal{G}}^s_{i,j}$ is responsible for detecting it. 
The receptive field of an output grid increases with $s$, so smaller output grids are better suited for detecting larger objects. 

The output grid cells $\hat{\mathcal{G}}^s_{i,j}$ contain $N_a$ anchor channels corresponding to anchor boxes $\mathbf{A}^s = \{(A_{w_a}, A_{h_a})\}_{a=1}^{N_a}$. A target object $\mathcal{O}$ is assigned to an anchor channel via tolerance-based matching of the object and anchor box sizes. This provides redundancy such that the grid cells $\hat{\mathcal{G}}^s_{i,j}$ can detect multiple objects and enables specialization for different object sizes and shapes. Additional detection redundancy is provided by also allowing the neighbouring grid cells $\hat{\mathcal{G}}^s_{i\pm1,j}$ and $\hat{\mathcal{G}}^s_{i,j\pm1}$ to detect an object in $\mathbf{I}_p$~\cite{wang2020scaled, glenn_jocher_2021_4679653}.

The $N_o$ output channels of $\hat{\mathcal{G}}^s_{i,j,a}$ contain the properties of a predicted object $\hat{\mathcal{O}}$, including the objectness $\hat{p}_{o}$ (the probability that an object exists), the intermediate bounding boxes $\hat{\mathbf{t}}'= (\hat{t}'_x, \hat{t}'_y, \hat{t}'_w, \hat{t}'_h)$, the object class scores $\mathbf{\hat{c}} = (\hat{c}_1, ..., \hat{c}_{K+1})$, and the intermediate keypoints $\hat{\mathbf{v}}'= \{(\hat{v}'_{xk}, \hat{v}'_{yk})\}_{k=1}^K$ for the human pose objects. Hence, $N_o = 3K + 6$.

\begin{figure}[t!]
\centering
    \includegraphics[trim={0cm, 0cm, 0cm, 0cm}, clip, width=0.37\linewidth]{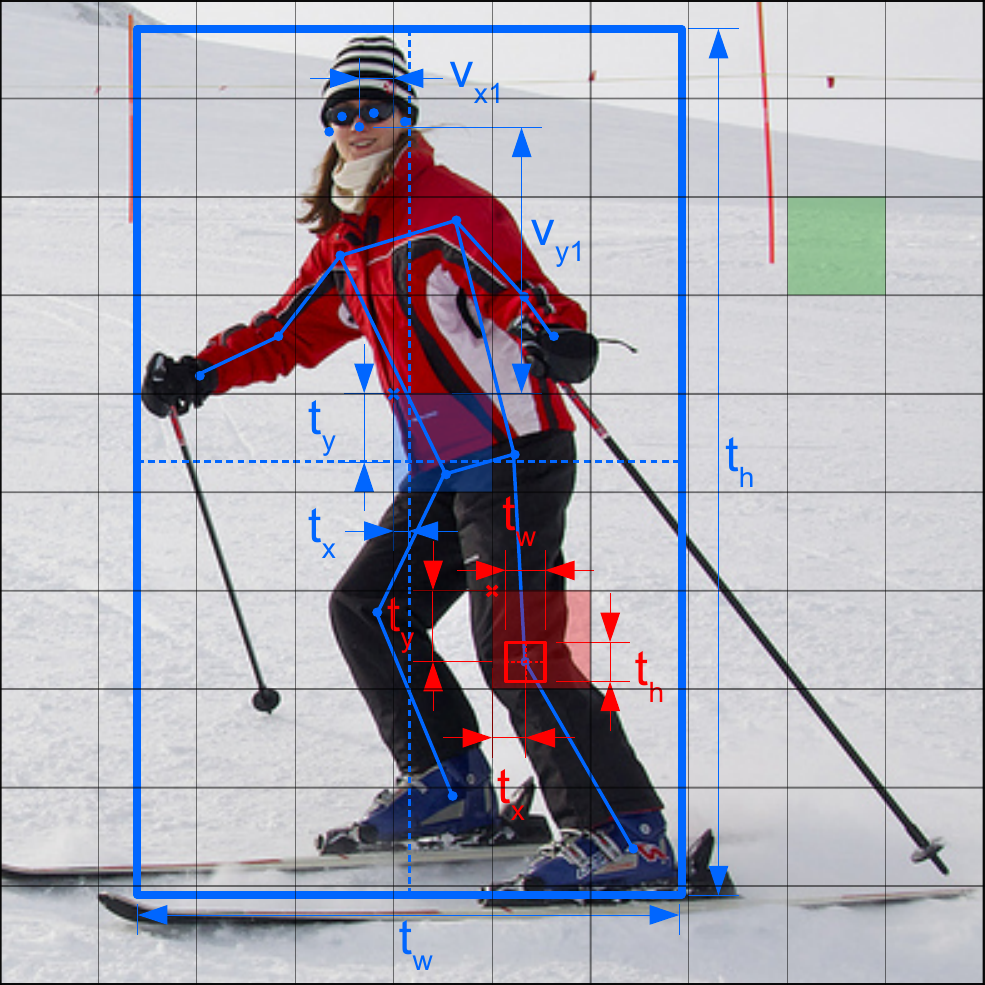}
    \includegraphics[trim={0cm, -1.3cm, 0cm, -0.5cm}, clip, width=0.62\linewidth]{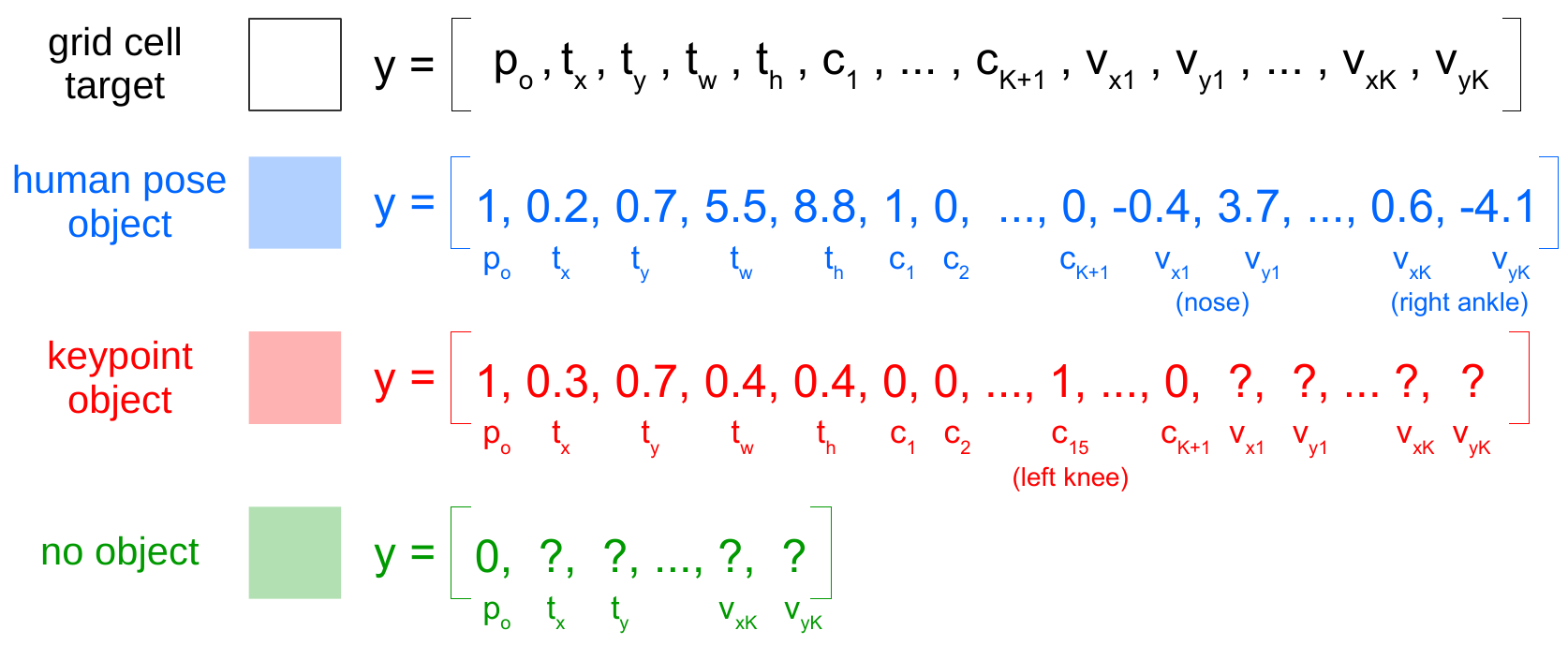}
\caption[Sample targets used for training KAPAO.]{Sample targets for training, including a human pose object (blue), keypoint object (red), and no object (green). The ``?'' values are not used in the loss computation.}
\label{fig:kapao_targets}
\end{figure}

Following~\cite{glenn_jocher_2021_4679653, wang2020scaled}, an object's intermediate bounding box $\hat{\mathbf{t}}$ is predicted in the grid coordinates and relative to the grid cell origin $(i, j)$ using:
\begin{equation}
    \hat{t}_x = 2\sigma(\hat{t}'_x) - 0.5 \quad\quad \hat{t}_y = 2\sigma(\hat{t}'_y) - 0.5
\end{equation}
\begin{equation}
    \hat{t}_w = \frac{A_w}{s}(2\sigma(\hat{t}'_w))^2 \quad\quad \hat{t}_h = \frac{A_h}{s}(2\sigma(\hat{t}'_h))^2.
\end{equation}
This detection strategy \hlll {is} extended to the keypoints of a pose object. A pose object's intermediate keypoints $\hat{\mathbf{v}}$ are predicted in the grid coordinates and relative to the grid cell origin $(i, j)$ using:
\begin{equation}
    \hat{v}_{xk} = \frac{A_w}{s}(4\sigma(\hat{v}'_{xk}) - 2) \quad\quad \hat{v}_{yk} = \frac{A_h}{s}(4\sigma(\hat{v}'_{yk}) - 2).
\end{equation}
The sigmoid function $\sigma$ facilitates learning by constraining the ranges of the object properties (\eg, $\hat{v}_{xk}$ and $\hat{v}_{yk}$ are constrained to $\pm2\frac{A_w}{s}$ and $\pm2\frac{A_h}{s}$, respectively). To learn $\hat{\mathbf{t}}$ and $\hat{\mathbf{v}}$, losses are applied in the grid space. Sample targets $\mathbf{t}$ and $\mathbf{v}$ are shown in Figure~\ref{fig:kapao_targets}.

\subsection{Loss Function}
\label{sec:kapao_loss}
A target set of grids $\mathbf{G}$ is constructed and a multi-task loss $\mathcal{L}(\hat{\mathbf{G}}, \mathbf{G})$ is applied to learn the objectness $\hat{p}_o$ ($\mathcal{L}_{obj})$, the intermediate bounding boxes $\hat{\textbf{t}}$ ($\mathcal{L}_{box}$), the class scores $\hat{\mathbf{c}}$ ($\mathcal{L}_{cls}$), and the intermediate pose object keypoints $\hat{\mathbf{v}}$ ($\mathcal{L}_{kps}$). The loss components are computed for a single image as follows:

\begin{equation}
    \mathcal{L}_{obj} = \sum_s \frac{\omega_s}{n{(G^s})}\sum_{G^s}\bce(\hat{p}_o, p_o\cdot \iou(\hat{\mathbf{t}}, \mathbf{t}))
\end{equation}
\begin{equation}
    \mathcal{L}_{box} = \sum_s \frac{1}{n{(\mathcal{O}\in G^s})}\sum_{\mathcal{O}\in G^s}1 - \iou(\hat{\mathbf{t}}, \mathbf{t})
\end{equation}
\begin{equation}
    \mathcal{L}_{cls} = \sum_s \frac{1}{n{(\mathcal{O}\in G^s})}\sum_{\mathcal{O}\in G^s}\bce(\hat{c}, c)
\end{equation}
\begin{equation}
    \mathcal{L}_{kps} = \sum_s \frac{1}{n{(\mathcal{O}^p\in G^s})}\sum_{\mathcal{O}^p\in G^s} \sum_{k=1}^K \delta(\nu_k > 0)||\hat{\mathbf{v}}_k - {\mathbf{v}}_k||_2
\end{equation}
where $\omega_s$ is the grid weighting, $\bce$ is the binary cross-entropy, $\iou$ is the complete intersection over union (CIoU)~\cite{zheng2020distance}, and $\nu_k$ are the keypoint visibility flags. When $\mathcal{G}^s_{i,j,a}$ represents a target object $\mathcal{O}$, the target objectness $p_o$ = 1 is multiplied by the $\iou$ to promote specialization amongst the anchor channel predictions~\cite{redmon2016you}. When $\mathcal{G}^s_{i,j,a}$ is not a target object, $p_o$ = 0. In practice, the losses are applied over a batch of images using batched grids. The total loss $\mathcal{L}$ is the weighted summation of the loss components scaled by the batch size $N_b$:
\begin{equation}
\mathcal{L} = N_b(\lambda_{obj}\mathcal{L}_{obj} + \lambda_{box}\mathcal{L}_{box} + \lambda_{cls}\mathcal{L}_{cls} + \lambda_{kps}\mathcal{L}_{kps}).
\end{equation}

\subsection{Inference}
The predicted intermediate bounding boxes $\hat{\mathbf{t}}$ and keypoints $\hat{\mathbf{v}}$ are mapped back to the original image coordinates using the following transformation:
\begin{equation}
    \hat{\mathbf{b}} = s(\hat{\mathbf{t}} + [i, j, 0, 0]) \quad\quad \hat{\mathbf{z}}_k = s(\hat{\mathbf{v}}_k + [i, j]).
\end{equation}
$\hat{\mathcal{G}}^s_{i,j,a}$ represents a positive pose object detection $\hat{\mathcal{O}}^p$ if its confidence $\hat{p}_o\cdot\max(\hat{\mathbf{c}})$ is greater than a threshold $\tau_{cp}$ and $\argmax(\hat{\mathbf{c}})=1$. Similarly, $\hat{\mathcal{G}}^s_{i,j,a}$ represents a positive keypoint object detection $\hat{\mathcal{O}}^k$ if $\hat{p}_o\cdot\max(\hat{\mathbf{c}}) > \tau_{ck}$ and $\argmax(\hat{\mathbf{c}}) > 1$, where the keypoint object class is $\argmax(\hat{\mathbf{c}}) - 1$. To remove redundant detections and obtain the candidate pose objects $\hat{\mathbf{O}}^{p\prime}$ and the candidate keypoint objects $\hat{\mathbf{O}}^{k\prime}$, the sets of positive pose object detections $\hat{\mathbf{O}}^p$ and positive keypoint object detections $\hat{\mathbf{O}}^p$ are filtered using non-maximum suppression (NMS) applied to the pose object bounding boxes with the $\iou$ thresholds $\tau_{bp}$ and $\tau_{bk}$:
\begin{equation}
    \hat{\mathbf{O}}^{p\prime} = \nms(\hat{\mathbf{O}}^p, \tau_{bp}) \quad\quad \hat{\mathbf{O}}^{k\prime} = \nms(\hat{\mathbf{O}}^k, \tau_{bk}).
\end{equation}
\hlll{It is noted that $\tau_{ck}$ and $\tau_{bk}$ are scalar thresholds used for all keypoint object classes.} Finally, the human pose predictions $\hat{\mathbf{P}} = \{\hat{\mathbf{P}}_i\in\mathbb{R}^{K \times 3}\}$ for $i \in \{1...n(\hat{\mathbf{O}}^{p\prime})\}$ are obtained by fusing the candidate keypoint objects with the candidate pose objects using a distance tolerance $\tau_{fd}$. To promote correct matches of keypoint objects to poses, the keypoint objects are only fused to pose objects with confidence $\hat{p}_o\cdot\max(\hat{\mathbf{c}}) > \tau_{fc}$:
\begin{equation}
    \hat{\mathbf{P}} = \varphi(\hat{\mathbf{O}}^{p\prime}, \hat{\mathbf{O}}^{k\prime}, \tau_{fd}, \tau_{fc}).
\end{equation}
The keypoint object fusion function $\varphi$ is defined in Algorithm~\ref{alg:fusion}, where the following notation is used to index an object's properties: $\hat{x} = \hat{\mathcal{O}}_x$ (\eg, a pose object's keypoints $\hat{\mathbf{z}}$ are referenced as $\hat{\mathcal{O}}^{p}_{\mathbf{z}}$).

\begin{algorithm}[t!]
\SetAlgoNoLine
\small
\DontPrintSemicolon
\newcommand\mycommfont[1]{\footnotesize\ttfamily\textcolor{blue}{#1}}
\SetCommentSty{mycommfont}

\SetKwInput{KwInput}{Input}                
\SetKwInput{KwOutput}{Output}              
\KwInput{$\hat{\mathbf{O}}^{p\prime}$, $\hat{\mathbf{O}}^{k\prime}$, $\tau_{fd}$, $\tau_{fc}$}
\KwOutput{$\hat{\mathbf{P}}$}

\If{$n(\hat{\mathbf{O}}^{p\prime}) > 0$}{
    $\hat{\mathbf{P}} \leftarrow \{0_{K \times 3} \mid \_\in\{1...n(\hat{\mathbf{O}}^{p\prime})\}\}$ \tcp{initialize poses}
    $\mathbf{\zeta} \leftarrow \{0 \mid \_\in\{1...n(\hat{\mathbf{O}}^{p\prime})\}\}$ \tcp{initialize pose confidences}
    
    \For{$(i, \hat{O}^p) \in \normalfont\textbf{enumerate}(\hat{\mathbf{O}}^{p\prime})$}{
    
        $\mathbf{\zeta}_i = \hat{O}^p_{p_o} \cdot\max(\hat{O}^p_\mathbf{c})$\;
        
        \For{$k\in\{1...K\}$}{
            $\hat{\mathbf{P}}_{i}[k] \leftarrow (\hat{\mathcal{O}}^p_{x_k}, \hat{\mathcal{O}}^p_{y_k}, 0)$ \tcp{assign pose object keypoints}
        }
    }
    $\hat{\mathbf{P}}^* \leftarrow \{\hat{\mathbf{P}}_{i} \in \hat{\mathbf{P}} \mid \mathbf{\zeta}_i > \tau_{fc}\}$ \tcp{poses above confidence threshold}
    
    \If{$n(\hat{\mathbf{P}}^*) > 0 \wedge n(\hat{\mathbf{O}}^{k\prime}) > 0$}{
    
        \For{$\hat{\mathcal{O}}^k \in \hat{\mathbf{O}}^{k\prime}$}{
        
            $k \leftarrow \argmax(\hat{\mathcal{O}}^k_\mathbf{c})-1$\ \tcp{keypoint index}
            $C_k \leftarrow \hat{\mathcal{O}}^k_{p_o}\max(\hat{\mathcal{O}}^k_\mathbf{c})$ \tcp{keypoint object confidence}
            
            $\mathbf{d}_i \leftarrow ||\hat{\mathbf{P}}_i^*[k, [1, 2]] - (\hat{\mathcal{O}}^k_{b_x}, \hat{\mathcal{O}}^k_{b_y})||_2$ \tcp{distances}
            
            $m \leftarrow \argmin({\mathbf{d}})$ \tcp{match index}
            
            \If{$\mathbf{d}_m < \tau_{fd} \wedge \hat{\mathbf{P}}_m^*[k, 3] < C_k$}{
            
                $\hat{\mathbf{P}}_m^{*}[k] = (\hat{\mathcal{O}}^k_{b_x}, \hat{\mathcal{O}}^k_{b_y}, C_k)$ \tcp{assign keypoint object to pose}
            }
        }
    }
}\Else{
    $\hat{\mathbf{P}}=\emptyset$ \tcp{empty set}
}
\caption{Keypoint object fusion ($\varphi$)}
\label{alg:fusion}
\end{algorithm}

\subsection{Limitations}
\label{sec:kapao_limitations}
A limitation of KAPAO is that pose objects do not include individual keypoint confidences, so the human pose predictions typically contain a sparse set of keypoint confidences $\hat{\mathbf{P}}_i[:,3]$ populated by the fused keypoint objects (see Algorithm~\ref{alg:fusion} for details). If desired, a complete set of keypoint confidences can be induced by only using keypoint objects, which is realized when $\tau_{ck} \rightarrow 0$. Another limitation is that training requires a considerable amount of time and GPU memory due to the large input size used. 

\section{Experiments}
\label{sec:kapao_exp}
We evaluate KAPAO on two multi-person human pose estimation datasets: COCO Keypoints~\cite{lin2014microsoft} \hlll{($K=17$)} and CrowdPose~\cite{li2019crowdpose} \hlll{($K=14$)}. We report the standard AP/AR detection metrics based on Object Keypoint Similarity \cite{lin2014microsoft} and compare against state-of-the-art methods.
All hyperparameters are provided in the source code.

\subsection{Microsoft COCO Keypoints}
\medskip\noindent\textbf{Training.} KAPAO-S/M/L were all trained for 500 epochs on COCO \texttt{train2017} using stochastic gradient descent with Nesterov momentum~\cite{Nesterov1983AMF}, weight decay, and a learning rate decayed over a single cosine cycle~\cite{loshchilov2016sgdr} with a 3-epoch warm-up period~\cite{goyal2017accurate}. The input images were resized and padded to 1280$\times$1280, keeping the original aspect ratio. Data augmentation used during training included mosaic~\cite{bochkovskiy2020yolov4}, HSV color-space perturbations, horizontal flipping, translations, and scaling. Many of the training hyperparameters were inherited from~\cite{wang2020scaled, glenn_jocher_2021_4679653}, including the anchor boxes $\mathbf{A}$ and the loss weights $w$, $\lambda_{obj}$, $\lambda_{box}$, and $\lambda_{cls}$. Others, including the keypoint bounding box size $b_s$ and the keypoint loss weight $\lambda_{kps}$, were manually tuned using a small grid search. The models were trained on four V100 GPUs with 32 GB memory each using batch sizes of 128, 72, and 48 for KAPAO-S, M, and L, respectively. Validation was performed after every epoch, saving the model weights that provided the highest validation AP.

\medskip\noindent\textbf{Testing.} The six inference parameters ($\tau_{cp}$, $\tau_{ck}$, $\tau_{bp}$, $\tau_{bk}$, $\tau_{fd}$, and $\tau_{fc}$) were manually tuned on the validation set using a coarse grid search to maximize accuracy. The results were not overly sensitive to the inference parameter values. When using TTA, the input image was scaled by factors of 0.8, 1, and 1.2, and the unscaled image was horizontally flipped. During post-processing, the multi-scale detections \hlll{were} concatenated before running NMS. When not using TTA, rectangular input images were used (\ie, 1280 px on the longest side), which marginally reduced the accuracy but increased the inference speed. 

\begin{table}[t]
\footnotesize
\centering
\begin{tabular}{l|c|c|c|cc|c|c|c}
	\hline
	Method & TTA & \makecell{Input Size(s)} & \makecell{Params\\(M)} & \makecell{FP\\(ms)} & \makecell{PP\\(ms)} & \makecell{Lat.\\(ms)} & AP & AR\\
	\hline
	HigherHRNet-W32~\cite{cheng2020higherhrnet} & N & 512 & 28.6 & 46.1 & 50.1 & 96.2 & 63.6 & 69.0\\
	\hll{\quad + SWAHR~\cite{luo2021rethinking}} & N & 512 & 28.6 & 45.1 & 86.6 & 132 & 64.7 & 70.3\\
	HigherHRNet-W32~\cite{cheng2020higherhrnet} & N & 640 & 28.6 & 52.4 & 71.4 & 124 & 64.9 & 70.3\\  
	HigherHRNet-W48~\cite{cheng2020higherhrnet} & N & 640 & 63.8 & 75.4 & 59.2 & 135 & 66.6 & 71.5\\
	\hll{\quad + SWAHR~\cite{luo2021rethinking}} & N & 640 & 63.8 & 86.3 & 194 & 280 & 67.3 & 73.0\\ 
	DEKR-W32~\cite{geng2021bottom} & N & 512 & 29.6 & 62.6 & 34.9 & 97.5 & 62.4 & 69.6\\ 
	DEKR-W48~\cite{geng2021bottom} & N & 640 & 65.7 & 109 & 45.8 & 155 & 66.3 & 73.2\\
	\hll{CenterGroup-W32~\cite{braso2021center}} & N & 512 & 30.3 & 98.9 & 16.0 & 115 & 66.9 & 71.6\\
	\hll{CenterGroup-W48~\cite{braso2021center}} & N & 640 & 65.5 & 155 & 14.5 & 170 & 69.1 & 74.0\\
	KAPAO-S & N & 1280 & \textbf{12.6} & \textbf{14.7} & \textbf{2.80} & \textbf{17.5} & 63.0 & 70.2\\  
	KAPAO-M & N & 1280 & 35.8  & 30.7 & 2.88 & 33.5 & 68.5 & 75.5 \\ 
	KAPAO-L & N & 1280 & 77.0  & 51.3 & 3.07 & 54.4 & \textbf{70.6} & \textbf{77.4} \\
	\hline
	HigherHRNet-W32~\cite{cheng2020higherhrnet} & Y & 256, 512, 1024 & 28.6 & 365 & 372 & 737 & 69.9 & 74.3\\  
	\hll{\quad + SWAHR~\cite{luo2021rethinking}} & Y & 256, 512, 1024 & 28.6 & 389 & 491 & 880 & 71.3 & 75.9\\
	HigherHRNet-W32~\cite{cheng2020higherhrnet} & Y & 320, 640, 1280 & 28.6 & 431 & 447 & 878 & 70.6 & 75.0\\  
	HigherHRNet-W48~\cite{cheng2020higherhrnet} & Y & 320, 640, 1280 & 63.8 & 643 & 436 & 1080 & 72.1 & 76.1\\  
	\hll{\quad + SWAHR~\cite{luo2021rethinking}} & Y & 320, 640, 1280 & 63.8 & 809 & 781 & 1590 & 73.0 & 77.6\\
	DEKR-W32~\cite{geng2021bottom} & Y & 256, 512, 1024 & 29.6 & 552 & 137 & 689 & 70.5 & 76.2\\  
	DEKR-W48~\cite{geng2021bottom} & Y & 320, 640, 1280 & 65.7 & 1010 & 157 & 1170 & 72.1 & 77.8\\
	\hll{CenterGroup-W32~\cite{braso2021center}} & Y & 256, 512, 1024 & 30.3 & 473 & 13.8 & 487 & 71.9 & 76.1\\
	\hll{CenterGroup-W48~\cite{braso2021center}} & Y & 320, 640, 1280 & 65.5 & 1050 & 11.8 & 1060 & \textbf{73.3} & 77.6\\
	KAPAO-S & Y & 1024, 1280, 1536 & \textbf{12.6} & \textbf{61.5} & \textbf{3.70} & \textbf{65.2} & 64.4 & 71.5\\  
	KAPAO-M & Y & 1024, 1280, 1536 & 35.8 & 126 & 3.67 & 130 & 69.9 & 76.8 \\
	KAPAO-L & Y & 1024, 1280, 1536 & 77.0 & 211 & 3.70 & 215 & 71.6 & \textbf{78.5} \\
	\hline
\end{tabular}
\smallskip
\caption{Accuracy and speed comparison with state-of-the-art single-stage human pose estimation models on COCO \texttt{val2017}, including the forward pass (FP) and post-processing (PP). Latencies (Lat.) averaged over \texttt{val2017} using a batch size of 1 on a TITAN Xp GPU.} 
\label{tab:kapao_coco_val}
\end{table}

\medskip\noindent\textbf{Results.} Table~\ref{tab:kapao_coco_val} compares the accuracy, forward pass (FP) time, and post-processing (PP) time of KAPAO with state-of-the-art single-stage methods HigherHRNet~\cite{cheng2020higherhrnet}, \hll{HigherHRNet + SWAHR~\cite{luo2021rethinking}, DEKR~\cite{geng2021bottom}, and CenterGroup~\cite{braso2021center}} on \texttt{val2017}. Two test settings were considered: (1) without any test-time augmentation (using a single forward pass of the network), and (2) with multi-scale and horizontal flipping test-time augmentation (TTA). It is noted that with the exception of CenterGroup, no inference speeds were reported in the original works. Rather, FLOPs were used as an indirect measure of computational efficiency. FLOPs are not only a poor indication of inference speed~\cite{ding2021repvgg}, but they are also only computed for the forward pass of the network and \hlll{thus} do not provide an indication of the amount of computation required for post-processing.

Due to expensive heatmap refinement, the post-processing times of HigherHRNet, HigherHRNet + SWAHR, and DEKR are at least an order of magnitude greater than KAPAO-L when not using TTA. The post-processing time of KAPAO depends less on the input size so it only increases by approximately 1 ms when using TTA. Conversely, HigherHRNet and HigherHRNet + SWAHR generate and refine large heatmaps during multi-scale testing and therefore require more than two orders of magnitude more post-processing time than KAPAO-L.

CenterGroup requires significantly less post-processing time than HigherHRNet and DEKR because it skips heatmap refinement and directly encodes pose center and keypoint heatmaps as embeddings that are fed to an attention-based grouping module. When not using TTA, CenterGroup-W48 provides an improvement of 2.5 AP over HigherHRNet-W48 and has a better accuracy-speed trade-off. \hlll{Still,} KAPAO-L is 3.1$\times$ faster than CenterGroup-W48 and 1.5 AP more accurate due to its efficient network architecture and near cost-free post-processing. When using TTA, KAPAO-L is \hlll{1.7 AP less accurate} than CenterGroup-W48, but 4.9$\times$ faster. KAPAO-L also achieves state-of-the-art AR, which is indicative of better detection rates. 

\hlll{We suspect that KAPAO is more accurate without TTA compared to previous methods because it uses larger input images; however, we emphasize that KAPAO consumes larger input sizes while still being faster than previous methods due to its well-designed network architecture and efficient post-processing. For the same reason, TTA (multi-scale testing in particular) doesn't provide as much of a benefit; input sizes $>$1280 are less effective due to the dataset images being limited to 640 px.} 

\begin{table}[t!]
\centering
\begin{tabular}{l|c|c|c|c|c|c|c}
	\hline
	Method & \hll{Lat. (ms)} & AP & AP$^{.50}$ & AP$^{.75}$ & AP$^M$ & AP$^L$ & AR\\
	\hline
	\hll{G-RMI~\cite{papandreou2017towards}$^\dagger$} & - & 64.9 & 85.5 & 71.3 & 62.3 & 70.0 & 69.7 \\
	\hll{RMPE~\cite{fang2017rmpe}$^\dagger$} & - & 61.8 & 83.7 & 69.8 & 58.6 & 67.6 & -\\
	CPN~\cite{chen2018cascaded}$^\dagger$ & - & 72.1 & 91.4 & 80.0 & 68.7 & 77.2 & 78.5 \\
	SimpleBaseline~\cite{xiao2018simple}$^\dagger$ & - & 73.7 & 91.9 & 81.1 & 70.3 & 80.0 & 79.0 \\
	HRNet-W48~\cite{sun2019deep}$^\dagger$ & - & 75.5 & \textbf{92.5} & \textbf{83.3} & 71.9 & \textbf{81.5} & 80.5\\
	EvoPose2D-L~\cite{mcnally2021evopose2d}$^\dagger$ & - & \textbf{75.7} & 91.9 & 83.1 & 72.2 & \textbf{81.5} & \textbf{81.7}\\
	MIPNet~\cite{khirodkar2021multi}$^\dagger$ & - & \textbf{75.7} & - & - & - & - & - \\
	RLE~\cite{li2021human}$^\dagger$ & - & \textbf{75.7} & 92.3 & 82.9 & \textbf{72.3} & 81.3 & - \\
	\hline
	OpenPose~\cite{cao2017realtime, cao2018openpose} & 74* & 61.8 & 84.9 & 67.5 & 57.1 & 68.2 & 66.5 \\
	Mask R-CNN~\cite{he2017mask} & - & 63.1 & 87.3 & 68.7 & 57.8 & 71.4 & - \\
	Associative Embeddings~\cite{newell2017associative} & - & 65.5 & 86.8 & 72.3 & 60.6 & 72.6 & 70.2\\
	PersonLab~\cite{papandreou2018personlab}  & - & 68.7 & 89.0 & 75.4 & 64.1 & 75.5 & 75.4 \\
	SPM~\cite{nie2019single} & - & 66.9 & 88.5 & 72.9 & 62.6 & 73.1 & - \\
	PifPaf~\cite{kreiss2019pifpaf} & - & 66.7 & - & - & 62.4 & 72.9 & - \\
	\hll{HGG~\cite{jin2020differentiable}} & - & 67.6 & 85.1 & 73.7 & 62.7 & 74.6 & 71.3 \\
	\hll{CenterNet~\cite{zhou2019objects}} & - & 63.0 & 86.8 & 69.6 & 58.9 & 70.4 & - \\
	\hll{Point-Set Anchors~\cite{wei2020point}} & - & 68.7 & 89.9 & 76.3 & 64.8 & 75.3 & - \\
	HigherHRNet-W48~\cite{cheng2020higherhrnet} & 1080 & 70.5 & 89.3 & 77.2 & 66.6 & 75.8 & 74.9\\ 
	\hll{\quad + SWAHR~\cite{luo2021rethinking}} & 1590 & \textbf{72.0} & 90.7 & \textbf{78.8} & \textbf{67.8} & \textbf{77.7} & - \\
	\hll{FCPose~\cite{mao2021fcpose}} & 93* & 65.6 & 87.9 & 72.6 & 62.1 & 72.3 & - \\
	DEKR-W48~\cite{geng2021bottom} & 1170 & 71.0 & 89.2 & 78.0 & 67.1 & 76.9 & 76.7\\
	\hll{CenterGroup-W48~\cite{braso2021center}} & 1060 & 71.4 & 90.5 & 78.1 & 67.2 & 77.5 & - \\
	KAPAO-S & \textbf{65.2} & 63.8 & 88.4 & 70.4 & 58.6 & 71.7 & 71.2 \\
	KAPAO-M & 130 & 68.8 & 90.5 & 76.5 & 64.3 & 76.0 & 76.3 \\
	KAPAO-L & 215 & 70.3 & \textbf{91.2} & 77.8 & 66.3 & 76.8 & \textbf{77.7} \\
	\hline
\end{tabular}
\smallskip
\caption[KAPAO compared to state-of-the-art two-stage and single-stage methods on COCO \texttt{test-dev}.]{Accuracy comparison with two-stage ($\dagger$) and single-stage methods on COCO \texttt{test-dev}. Best results \hlll{reported} (\ie, including TTA). DEKR results use a model-agnostic rescoring network~\cite{geng2021bottom}. \hll{Latencies (Lat.) taken from Table~\ref{tab:kapao_coco_val}. *Latencies reported in original papers~\cite{cao2018openpose, mao2021fcpose} and measured using an NVIDIA GTX 1080Ti GPU.}}
\label{tab:kapao_coco_test}
\end{table}

In Table~\ref{tab:kapao_coco_test}, the accuracy of KAPAO is compared to single-stage and two-stage methods on \texttt{test-dev}. KAPAO-L achieves state-of-the-art AR and falls within \hll{1.7 AP of \hlll{the} best performing single-stage method HigherHRNet-W48 + SWAHR while being 7.4$\times$ faster. Notably, KAPAO-L is more accurate than the early two-stage methods G-RMI~\cite{papandreou2017towards} and RMPE~\cite{fang2017rmpe} and \hlll{popular} single-stage methods like OpenPose~\cite{cao2017realtime, cao2018openpose}, Associative Embeddings~\cite{newell2017associative}, and PersonLab~\cite{papandreou2018personlab}. Compared to other single-stage methods that extend object detectors for human pose estimation (Mask R-CNN~\cite{he2017mask}, CenterNet~\cite{zhou2019objects}, Point-Set Anchors~\cite{wei2020point}, and FCPose~\cite{mao2021fcpose}), KAPAO-L is considerably more accurate. Among all the single-stage methods, KAPAO-L achieves state-of-the-art AP at an OKS threshold of 0.50}, which is indicative of better detection rates but less precise keypoint localization. This is an area to explore in future work.

\subsection{CrowdPose}
\label{sec:kapao_crowdpose}
KAPAO was trained on the \texttt{trainval} split with 12k images and was evaluated on the 8k images in \texttt{test}. The same training and inference settings as on COCO were used except the models were trained for 300 epochs and no validation was performed \hlll{during training}. The final model weights were used for testing. Table~\ref{tab:kapao_crowdpose_test} compares the accuracy of KAPAO against state-of-the-art methods. It was found that KAPAO excels in the presence of occlusion, \hll{achieving competitive results across all metrics compared to previous single-stage methods and state-of-the-art accuracy for AP$^{.50}$.} The proficiency of KAPAO in crowded scenes is clear when analyzing AP$^E$, AP$^M$, and AP$^H$: KAPAO-L and DEKR-W48~\cite{geng2021bottom} perform equally on images with easy Crowd Index (less occlusion), but KAPAO-L is 1.1 AP more accurate for both medium and hard Crowd Indices (more occlusion). 

\begin{table}[t!]
\footnotesize
\centering
\begin{tabular}{l|c|c|c|c|c|c|c}
	\hline
	Method & \hll{Lat. (ms)} & AP & AP$^{.50}$ & AP$^{.75}$ & AP$^E$ & AP$^M$ & AP$^H$\\
	\hline
	Mask R-CNN~\cite{he2017mask} & - & 57.2 & 83.5 & 60.3 & 69.4 & 57.9 & 45.8 \\
	AlphaPose~\cite{fang2017rmpe}$^\dagger$ & - & 61.0 & 81.3 & \textbf{66.0} & \textbf{71.2} & \textbf{61.4} & \textbf{51.1} \\
	\hll{SimpleBaseline~\cite{xiao2018simple}}$^\dagger$ & - & 60.8 & 81.4 & 65.7 & 71.4 & 61.2 & 51.2 \\
	SPPE~\cite{li2019crowdpose} & - & 66.0 & 84.2 & 71.5 & 75.5 & 66.3 & 57.4 \\
	MIPNet~\cite{khirodkar2021multi}$^\dagger$ & - & \textbf{70.0} & - & - & - & - & - \\
	\hline
	OpenPose~\cite{cao2017realtime} & 74* & - & - & - & 62.7 & 48.7 & 32.3 \\
	HigherHRNet-W48~\cite{cheng2020higherhrnet} & 1080 & 67.6 & 87.4 & 72.6 & 75.8 & 68.1 & 58.9\\
	DEKR-W48~\cite{geng2021bottom} & 1170 & 68.0 & 85.5 & 73.4 & 76.6 & 68.8 & 58.4\\
	\hll{CenterGroup-W48~\cite{braso2021center}} & 1060 & \textbf{70.0} & 88.9 & \textbf{75.7} & \textbf{77.3} & \textbf{70.8} & \textbf{63.2} \\
	KAPAO-S & 65.2 & 63.8 & 87.7 & 69.4 & 72.1 & 64.8 & 53.2 \\
	KAPAO-M & 130 & 67.1 & 88.8 & 73.4 & 75.2 & 68.1 & 56.9 \\
	KAPAO-L & 215 & 68.9 & \textbf{89.4} & 75.6 & 76.6 & 69.9 & 59.5 \\
	\hline
\end{tabular}
\smallskip
\caption[KAPAO compared to state-of-the-art two-stage and single-stage methods on CrowdPose.]{Comparison with single-stage and two-stage ($\dagger$) methods on CrowdPose \texttt{test}, including TTA. DEKR results use a model-agnostic rescoring network~\cite{geng2021bottom}. \hll{HigherHRNet + SWAHR~\cite{luo2021rethinking} not included due to issues reproducing the results reported in the paper using the source code. Latencies (Lat.) taken from Table~\ref{tab:kapao_coco_val}. *Latency reported in original paper~\cite{cao2018openpose} and measured using NVIDIA GTX 1080Ti GPU on COCO.}}
\label{tab:kapao_crowdpose_test}
\end{table}

\subsection{Ablation Studies}
\label{sec:kapao_ablation}

\begin{figure}[t!]
\centering
    \includegraphics[trim={0cm, 0cm, 0cm, 0cm}, clip, width=0.495\linewidth]{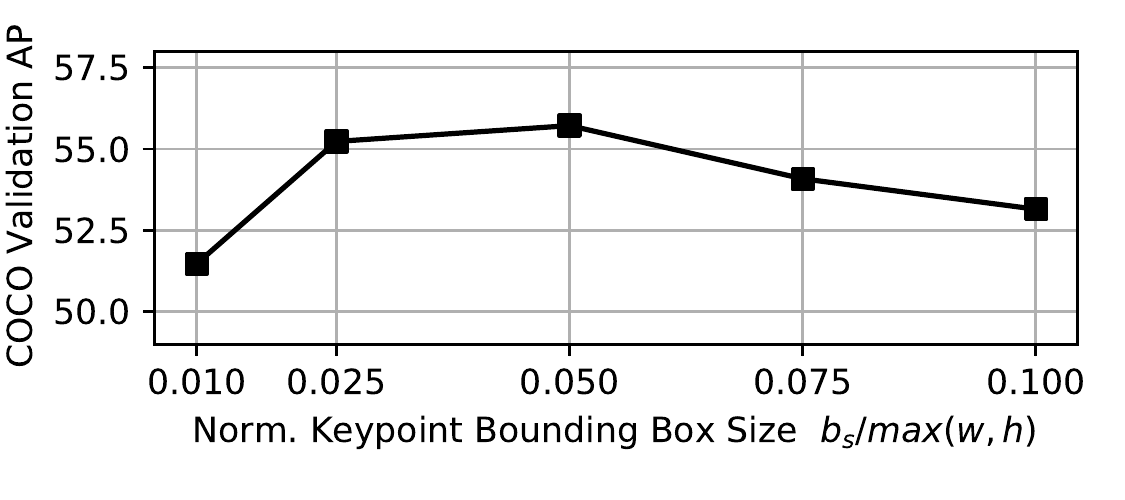}
    \includegraphics[trim={0cm, 0cm, 0cm, 0cm}, clip, width=0.495\linewidth]{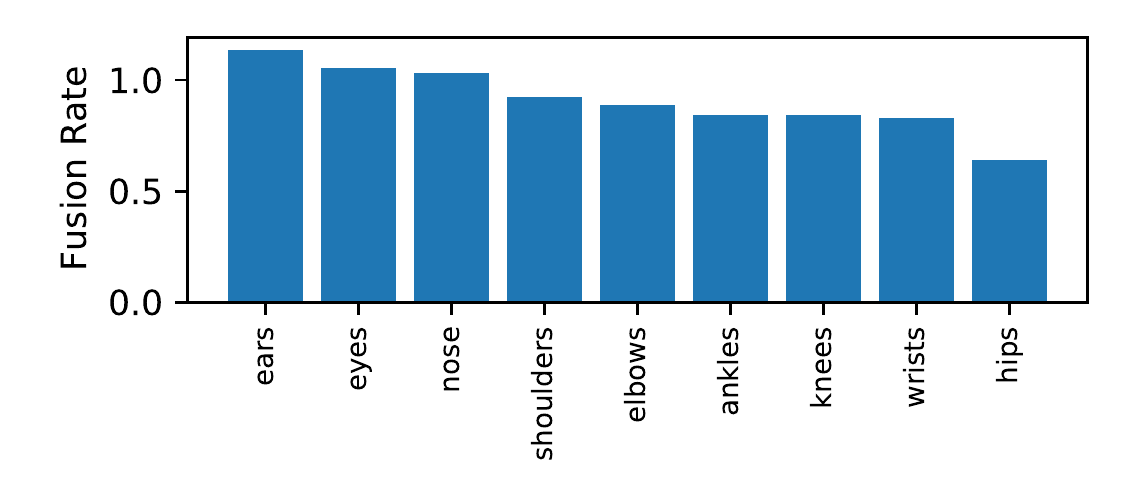}
\caption{Left: The influence of keypoint object bounding box size on learning. Each KAPAO-S model was trained for 50 epochs. Right: Keypoint object fusion rates for each keypoint type. Evaluated on COCO \texttt{val2017} using KAPAO-S without TTA.}
\label{fig:kapao_kp_bbox_fusion}
\end{figure}

The influence of the keypoint bounding box size $b_s$, one of KAPAO's important hyperparameters, was empirically analyzed. Five KAPAO-S models were trained on COCO \texttt{train2017} for 50 epochs using normalized keypoint bounding box sizes $b_s/max(w,h)$ $\in\{0.01, 0.025, 0.05, 0.075, 0.1\}$. The validation AP is plotted in Figure~\ref{fig:kapao_kp_bbox_fusion} (left). The results are consistent with the prior work of McNally \etal~\cite{mcnally2021deepdarts}: $b_s/max(w,h) <$ 2.5\% destabilizes training leading to poor accuracy, and optimal $b_s/max(w,h)$ is observed around 5\% (used for the experiments in previous section). In contrast to McNally \etal, the accuracy in this study degrades quickly for $b_s/max(w,h) >$ 5\%. It is hypothesized that large $b_s$ in this application interferes with pose object learning. 

The accuracy improvements resulting from fusing the keypoint objects with the pose objects are provided in Table~\ref{tab:kapao_det_fusion}. Keypoint object fusion adds no less than 1.0 AP and over 3.0 AP in some cases. Moreover, keypoint object fusion is fast; the added post-processing time per image is $\leq$ 1.7 ms on COCO and $\leq$ 4.5 ms on CrowdPose. Relative to the time required for the forward pass of the network (see Table~\ref{tab:kapao_coco_val}), these are small increases.

\begin{table}[t]
\footnotesize
\centering
\begin{tabular}{l|c|c|c}
	\hline
	Method & TTA & \makecell{$\Delta$ Lat. (ms) / $\Delta$AP\\(COCO \texttt{val2017})} & \makecell{$\Delta$ Lat. (ms) / $\Delta$AP\\(CrowdPose \texttt{test})}\\
	\hline
	KAPAO-S & N & +1.2 / +2.4 & +3.3 / +2.9 \\
	KAPAO-M & N & +1.2 / +1.1 & +3.5 / +3.2\\
	KAPAO-L & N & +1.7 / +1.2 & +4.2 / +1.0 \\
    \hline
	KAPAO-S & Y & +1.7 / +2.8 & +3.9 / +3.2 \\
	KAPAO-M & Y & +1.6 / +1.5 & +4.4 / +3.5 \\
	KAPAO-L & Y & +1.4 / +1.1 & +4.5 / +1.0 \\
	\hline
\end{tabular}
\smallskip
\caption[Accuracy improvement resulting from fusing keypoint objects with pose objects.]{Accuracy improvement when fusing keypoint object detections with human pose detections. Latencies averaged over each dataset using a batch size of 1 on a TITAN Xp GPU.}
\label{tab:kapao_det_fusion}
\end{table}

The fusion of keypoint objects by class is also studied. Figure~\ref{fig:kapao_kp_bbox_fusion} (right) plots the fusion rates for each keypoint type for KAPAO-S with no TTA on COCO \texttt{val2017}. The fusion rate is equal to the number of fused keypoint objects divided by the number of keypoints of that type in the dataset. Because the number of human pose predictions is generally greater than the actual number of person instances in the dataset, the fusion rate can be greater than 1. As originally hypothesized, keypoints that are characterized by distinct local image features (\eg, the eyes, ears, and nose) have higher fusion rates as they are detected more precisely as keypoint objects than as pose objects. Conversely, keypoints that require a more global understanding (\eg, the hips) are better detected using pose objects, as evidenced by lower fusion rates.

\section{Conclusion}
This paper presents KAPAO, a heatmap-free keypoint estimation method based on modeling keypoints and poses as objects. KAPAO is effectively applied to the problem of single-stage multi-person human pose estimation by detecting human pose objects. Moreover, fusing jointly detected keypoint objects improves the accuracy of the predicted human poses with minimal computational overhead. When not using test-time augmentation, KAPAO is significantly faster and more accurate than previous single-stage methods, which are impeded greatly by heatmap post-processing and bottom-up keypoint grouping. Moreover, KAPAO performs well in the presence of heavy occlusion as evidenced by competitive results on CrowdPose.

{\bigskip\noindent\small\textbf{Acknowledgements.} This work was supported in part by Compute Canada, the Canada Research Chairs Program, the Natural Sciences and Engineering Research Council of Canada, a Microsoft Azure Grant, and an NVIDIA Hardware Grant.}

\bibliographystyle{splncs04}
\bibliography{egbib}

\begin{thebibliography}{10}
\providecommand{\url}[1]{\texttt{#1}}
\providecommand{\urlprefix}{URL }
\providecommand{\doi}[1]{https://doi.org/#1}

\bibitem{andriluka2018posetrack}
Andriluka, M., Iqbal, U., Insafutdinov, E., Pishchulin, L., Milan, A., Gall,
  J., Schiele, B.: Posetrack: A benchmark for human pose estimation and
  tracking. In: CVPR (2018)

\bibitem{bochkovskiy2020yolov4}
Bochkovskiy, A., Wang, C.Y., Liao, H.Y.M.: Yolov4: Optimal speed and accuracy
  of object detection. arXiv preprint arXiv:2004.10934  (2020)

\bibitem{braso2021center}
Bras{\'o}, G., Kister, N., Leal-Taix{\'e}, L.: The center of attention:
  Center-keypoint grouping via attention for multi-person pose estimation. In:
  ICCV (2021)

\bibitem{cao2018openpose}
Cao, Z., Hidalgo, G., Simon, T., Wei, S.E., Sheikh, Y.: Openpose: realtime
  multi-person 2d pose estimation using part affinity fields. arXiv preprint
  arXiv:1812.08008  (2018)

\bibitem{cao2017realtime}
Cao, Z., Simon, T., Wei, S.E., Sheikh, Y.: Realtime multi-person 2d pose
  estimation using part affinity fields. In: CVPR (2017)

\bibitem{chen2018cascaded}
Chen, Y., Wang, Z., Peng, Y., Zhang, Z., Yu, G., Sun, J.: Cascaded pyramid
  network for multi-person pose estimation. In: CVPR (2018)

\bibitem{cheng2020higherhrnet}
Cheng, B., Xiao, B., Wang, J., Shi, H., Huang, T.S., Zhang, L.: {HigherHRNet}:
  Scale-aware representation learning for bottom-up human pose estimation. In:
  CVPR (2020)

\bibitem{ding2021repvgg}
Ding, X., Zhang, X., Ma, N., Han, J., Ding, G., Sun, J.: Rep{VGG}: Making
  {VGG}-style convnets great again. In: CVPR (2021)

\bibitem{dong2018style}
Dong, X., Yan, Y., Ouyang, W., Yang, Y.: Style aggregated network for facial
  landmark detection. In: CVPR (2018)

\bibitem{fang2017rmpe}
Fang, H.S., Xie, S., Tai, Y.W., Lu, C.: {RMPE}: Regional multi-person pose
  estimation. In: ICCV (2017)

\bibitem{gavrilyuk2020actor}
Gavrilyuk, K., Sanford, R., Javan, M., Snoek, C.G.: Actor-transformers for
  group activity recognition. In: CVPR (2020)

\bibitem{geng2021bottom}
Geng, Z., Sun, K., Xiao, B., Zhang, Z., Wang, J.: Bottom-up human pose
  estimation via disentangled keypoint regression. In: CVPR (2021)

\bibitem{goyal2017accurate}
Goyal, P., Doll{\'a}r, P., Girshick, R., Noordhuis, P., Wesolowski, L., Kyrola,
  A., Tulloch, A., Jia, Y., He, K.: Accurate, large minibatch sgd: Training
  imagenet in 1 hour. arXiv preprint arXiv:1706.02677  (2017)

\bibitem{he2017mask}
He, K., Gkioxari, G., Doll{\'a}r, P., Girshick, R.: Mask {R-CNN}. In: ICCV
  (2017)

\bibitem{huang2020awr}
Huang, W., Ren, P., Wang, J., Qi, Q., Sun, H.: Awr: Adaptive weighting
  regression for 3d hand pose estimation. In: AAAI (2020)

\bibitem{insafutdinov2016deepercut}
Insafutdinov, E., Pishchulin, L., Andres, B., Andriluka, M., Schiele, B.:
  {DeeperCut}: A deeper, stronger, and faster multi-person pose estimation
  model. In: ECCV (2016)

\bibitem{iqbal2016multi}
Iqbal, U., Gall, J.: Multi-person pose estimation with local joint-to-person
  associations. In: ECCV (2016)

\bibitem{iqbal2018hand}
Iqbal, U., Molchanov, P., Breuel Juergen~Gall, T., Kautz, J.: Hand pose
  estimation via latent 2.5 d heatmap regression. In: ECCV (2018)

\bibitem{jakab2018unsupervised}
Jakab, T., Gupta, A., Bilen, H., Vedaldi, A.: Unsupervised learning of object
  landmarks through conditional image generation. In: NeurIPS (2018)

\bibitem{jeon2019joint}
Jeon, S., Min, D., Kim, S., Sohn, K.: Joint learning of semantic alignment and
  object landmark detection. In: ICCV (2019)

\bibitem{jia2016dynamic}
Jia, X., De~Brabandere, B., Tuytelaars, T., Gool, L.V.: Dynamic filter
  networks. NeurIPS  (2016)

\bibitem{jin2020differentiable}
Jin, S., Liu, W., Xie, E., Wang, W., Qian, C., Ouyang, W., Luo, P.:
  Differentiable hierarchical graph grouping for multi-person pose estimation.
  In: ECCV (2020)

\bibitem{glenn_jocher_2021_4679653}
Jocher, G., {et. al.}: {ultralytics/yolov5: v5.0} (Apr 2021).
  \doi{10.5281/zenodo.4679653}

\bibitem{khirodkar2021multi}
Khirodkar, R., Chari, V., Agrawal, A., Tyagi, A.: Multi-hypothesis pose
  networks: Rethinking top-down pose estimation. In: ICCV (2021)

\bibitem{kreiss2019pifpaf}
Kreiss, S., Bertoni, L., Alahi, A.: Pifpaf: Composite fields for human pose
  estimation. In: CVPR (2019)

\bibitem{lecun1995convolutional}
LeCun, Y., Bengio, Y., et~al.: Convolutional networks for images, speech, and
  time series. The handbook of brain theory and neural networks
  \textbf{3361}(10) (1995)

\bibitem{li2021human}
Li, J., Bian, S., Zeng, A., Wang, C., Pang, B., Liu, W., Lu, C.: Human pose
  regression with residual log-likelihood estimation. In: ICCV (2021)

\bibitem{li2019crowdpose}
Li, J., Wang, C., Zhu, H., Mao, Y., Fang, H.S., Lu, C.: Crowdpose: Efficient
  crowded scenes pose estimation and a new benchmark. In: CVPR (2019)

\bibitem{li2021pose}
Li, K., Wang, S., Zhang, X., Xu, Y., Xu, W., Tu, Z.: Pose recognition with
  cascade transformers. In: CVPR (2021)

\bibitem{li20212d}
Li, Y., Yang, S., Zhang, S., Wang, Z., Yang, W., Xia, S.T., Zhou, E.: Is 2d
  heatmap representation even necessary for human pose estimation? arXiv
  preprint arXiv:2107.03332  (2021)

\bibitem{lin2017feature}
Lin, T.Y., Doll{\'a}r, P., Girshick, R., He, K., Hariharan, B., Belongie, S.:
  Feature pyramid networks for object detection. In: CVPR (2017)

\bibitem{lin2017focal}
Lin, T.Y., Goyal, P., Girshick, R., He, K., Doll{\'a}r, P.: Focal loss for
  dense object detection. In: ICCV (2017)

\bibitem{lin2014microsoft}
Lin, T.Y., Maire, M., Belongie, S., Hays, J., Perona, P., Ramanan, D.,
  Doll{\'a}r, P., Zitnick, C.L.: Microsoft {COCO}: Common objects in context.
  In: ECCV (2014)

\bibitem{loshchilov2016sgdr}
Loshchilov, I., Hutter, F.: {SGDR}: Stochastic gradient descent with warm
  restarts. In: ICLR (2017)

\bibitem{luo2021rethinking}
Luo, Z., Wang, Z., Huang, Y., Wang, L., Tan, T., Zhou, E.: Rethinking the
  heatmap regression for bottom-up human pose estimation. In: CVPR (2021)

\bibitem{mao2021fcpose}
Mao, W., Tian, Z., Wang, X., Shen, C.: Fcpose: Fully convolutional multi-person
  pose estimation with dynamic instance-aware convolutions. In: CVPR (2021)

\bibitem{mcnally2021evopose2d}
McNally, W., Vats, K., Wong, A., McPhee, J.: {EvoPose2D}: Pushing the
  boundaries of 2d human pose estimation using accelerated neuroevolution with
  weight transfer. IEEE Access  (2021)

\bibitem{mcnally2021deepdarts}
McNally, W., Walters, P., Vats, K., Wong, A., McPhee, J.: {DeepDarts}: Modeling
  keypoints as objects for automatic scorekeeping in darts using a single
  camera. In: CVPRW (2021)

\bibitem{mcnally2018action}
McNally, W., Wong, A., McPhee, J.: Action recognition using deep convolutional
  neural networks and compressed spatio-temporal pose encodings. Journal of
  Computational Vision and Imaging Systems  \textbf{4}(1), ~3--3 (2018)

\bibitem{mcnally2019star}
McNally, W., Wong, A., McPhee, J.: {STAR-Net}: Action recognition using
  spatio-temporal activation reprojection. In: CRV (2019)

\bibitem{Nesterov1983AMF}
Nesterov, Y.: A method for solving the convex programming problem with
  convergence rate o(1/k2). Proceedings of the USSR Academy of Sciences
  \textbf{269},  543--547 (1983)

\bibitem{newell2017associative}
Newell, A., Huang, Z., Deng, J.: Associative embedding: End-to-end learning for
  joint detection and grouping. In: NeurIPS (2017)

\bibitem{newell2016stacked}
Newell, A., Yang, K., Deng, J.: Stacked hourglass networks for human pose
  estimation. In: ECCV (2016)

\bibitem{nie2019single}
Nie, X., Feng, J., Zhang, J., Yan, S.: Single-stage multi-person pose machines.
  In: ICCV (2019)

\bibitem{papandreou2018personlab}
Papandreou, G., Zhu, T., Chen, L.C., Gidaris, S., Tompson, J., Murphy, K.:
  Personlab: Person pose estimation and instance segmentation with a bottom-up,
  part-based, geometric embedding model. In: ECCV (2018)

\bibitem{papandreou2017towards}
Papandreou, G., Zhu, T., Kanazawa, N., Toshev, A., Tompson, J., Bregler, C.,
  Murphy, K.: Towards accurate multi-person pose estimation in the wild. In:
  CVPR (2017)

\bibitem{pavllo20193d}
Pavllo, D., Feichtenhofer, C., Grangier, D., Auli, M.: 3d human pose estimation
  in video with temporal convolutions and semi-supervised training. In: CVPR
  (2019)

\bibitem{pishchulin2016deepcut}
Pishchulin, L., Insafutdinov, E., Tang, S., Andres, B., Andriluka, M., Gehler,
  P.V., Schiele, B.: {DeepCut}: Joint subset partition and labeling for multi
  person pose estimation. In: CVPR (2016)

\bibitem{raaj2019efficient}
Raaj, Y., Idrees, H., Hidalgo, G., Sheikh, Y.: Efficient online multi-person 2d
  pose tracking with recurrent spatio-temporal affinity fields. In: CVPR (2019)

\bibitem{redmon2016you}
Redmon, J., Divvala, S., Girshick, R., Farhadi, A.: You only look once:
  Unified, real-time object detection. In: CVPR (2016)

\bibitem{redmon2018yolov3}
Redmon, J., Farhadi, A.: Yolov3: An incremental improvement. arXiv preprint
  arXiv:1804.02767  (2018)

\bibitem{ren2015faster}
Ren, S., He, K., Girshick, R., Sun, J.: {Faster R-CNN}: Towards real-time
  object detection with region proposal networks (2015)

\bibitem{rezende2015variational}
Rezende, D., Mohamed, S.: Variational inference with normalizing flows. In:
  ICML (2015)

\bibitem{sun2019deep}
Sun, K., Xiao, B., Liu, D., Wang, J.: Deep high-resolution representation
  learning for human pose estimation. In: CVPR (2019)

\bibitem{suwajanakorn2018discovery}
Suwajanakorn, S., Snavely, N., Tompson, J., Norouzi, M.: Discovery of latent 3d
  keypoints via end-to-end geometric reasoning. In: NeurIPS (2018)

\bibitem{tian2019fcos}
Tian, Z., Shen, C., Chen, H., He, T.: {FCOS}: Fully convolutional one-stage
  object detection. In: ICCV (2019)

\bibitem{tompson2014joint}
Tompson, J.J., Jain, A., LeCun, Y., Bregler, C.: Joint training of a
  convolutional network and a graphical model for human pose estimation. In:
  NeurIPS (2014)

\bibitem{toshev2014deeppose}
Toshev, A., Szegedy, C.: {DeepPose}: Human pose estimation via deep neural
  networks. In: CVPR (2014)

\bibitem{vats2019pucknet}
Vats, K., McNally, W., Dulhanty, C., Lin, Z.Q., Clausi, D.A., Zelek, J.:
  {PuckNet}: Estimating hockey puck location from broadcast video. In: AAAI
  Workshops (2019)

\bibitem{voeikov2020ttnet}
Voeikov, R., Falaleev, N., Baikulov, R.: Ttnet: Real-time temporal and spatial
  video analysis of table tennis. In: CVPRW (2020)

\bibitem{wang2020scaled}
Wang, C.Y., Bochkovskiy, A., Liao, H.Y.M.: {Scaled-YOLOv4}: Scaling cross stage
  partial network. arXiv preprint arXiv:2011.08036  (2020)

\bibitem{wang2020cspnet}
Wang, C.Y., Liao, H.Y.M., Wu, Y.H., Chen, P.Y., Hsieh, J.W., Yeh, I.H.: Cspnet:
  A new backbone that can enhance learning capability of cnn. In: CVPR (2020)

\bibitem{wang2019adaptive}
Wang, X., Bo, L., Fuxin, L.: Adaptive wing loss for robust face alignment via
  heatmap regression. In: ICCV (2019)

\bibitem{wei2020point}
Wei, F., Sun, X., Li, H., Wang, J., Lin, S.: Point-set anchors for object
  detection, instance segmentation and pose estimation. In: ECCV (2020)

\bibitem{wei2016convolutional}
Wei, S.E., Ramakrishna, V., Kanade, T., Sheikh, Y.: Convolutional pose
  machines. In: CVPR (2016)

\bibitem{xiao2018simple}
Xiao, B., Wu, H., Wei, Y.: Simple baselines for human pose estimation and
  tracking. In: ECCV (2018)

\bibitem{xu2021anchorface}
Xu, Z., Li, B., Yuan, Y., Geng, M.: {AnchorFace}: An anchor-based facial
  landmark detector across large poses. In: AAAI (2021)

\bibitem{yang2021transpose}
Yang, S., Quan, Z., Nie, M., Yang, W.: Transpose: Keypoint localization via
  transformer. In: ICCV (2021)

\bibitem{zheng2020distance}
Zheng, Z., Wang, P., Liu, W., Li, J., Ye, R., Ren, D.: {Distance-IoU loss}:
  Faster and better learning for bounding box regression. In: AAAI (2020)

\bibitem{zhou2019objects}
Zhou, X., Wang, D., Kr{\"a}henb{\"u}hl, P.: Objects as points. arXiv preprint
  arXiv:1904.07850  (2019)

\end{thebibliography}


\begin{thebibliography}{2}
\bibitem[A1]{app_malawski2016classification}
Malawski, F., Kwolek, B.: Classification of basic footwork in fencing using accelerometer. In: Signal Processing: Algorithms, Architectures, Arrangements, and
Applications (SPA) (2016)

\bibitem[A2]{app_zhu2022fencenet}
Zhu, K., McPhee, J., Wong, A.: FenceNet: Fine-grained Footwork Prediction in Fencing. Submitted to CVSports (2022)

\bibitem[A3]{app_ronchi2017benchmarking}
Ronchi, M.R., Perona, P.: Benchmarking and error diagnosis in multi-instance pose estimation. In: ICCV (2017)
\end{thebibliography}


\clearpage
\appendix

\section{Supplementary Material}

\subsection{Hyperparameters}
For convenience, the KAPAO hyperparameters used to generate the results in this paper are provided in Table~\ref{tab:hyp} in the order they appear in the text. Other hyperparameters not referenced in the text (\eg, augmentation settings) are included in the code. Many of the hyperparameters are inherited from~\cite{glenn_jocher_2021_4679653}, where an evolutionary algorithm was used to search for optimal values for object detection on COCO. Some hyperparameters, such as the keypoint bounding box size $b_s$ and the keypoint loss weight $\lambda_{kps}$, were manually tuned using a small grid search. The influence of $b_s$ is studied in Section~\ref{sec:kapao_ablation}. Relative to KAPAO-L, the number of layers and channels in KAPAO-M were scaled by 2/3 and 3/4, respectively. Similarly, the number of layers and channels in KAPAO-S were scaled by 1/3 and 1/2, respectively.

\begin{table}[h]
\centering
\begin{tabular}{l|c|c}
    \hline
    \hll{Hyperparameter Description} & Symbol & Value(s)\\
    \hline
    output grid scales & $s$ & $\{8, 16, 32, 64\}$\\
    keypoint object bounding box size (px) & $b_s$ & $64$\\
    input image height, width (px) & $h$, $w$ & $1280$, $1280$\\
    $\mathcal{G}^8$ anchor boxes (width, height) (px) & $\mathbf{A}^8$ & $\{(19, 27), (44, 40), (38, 94)\}$\\
    $\mathcal{G}^{16}$ anchor boxes (width, height) (px) & $\mathbf{A}^{16}$ & $\{(96, 68), (86, 152), (180, 137)\}$\\
    $\mathcal{G}^{32}$ anchor boxes (width, height) (px) & $\mathbf{A}^{32}$ & $\{(140, 301), (303, 264), (238, 542)\}$\\
    $\mathcal{G}^{64}$ anchor boxes (width, height) (px) & $\mathbf{A}^{64}$ & $\{(436, 615), (739, 380), (925, 792)\}$\\
    loss weights for $\mathcal{G}^8$, $\mathcal{G}^{16}$ $\mathcal{G}^{32}$, and $\mathcal{G}^{64}$ & $\omega$ & $\{4.0, 1.0, 0.25, 0.06\}$\\
    objectness loss weight & $\lambda_{obj}$ & $0.7 \times (w/640)^2 \times 3/n(s)$\\
    bounding box loss weight & $\lambda_{box}$ & $0.05 \times 3/n(s)$\\
    class loss weight & $\lambda_{cls}$ & $0.3 \times (K+1)/80 \times 3/n(s)$\\
    pose object keypoints loss weight & $\lambda_{kps}$ & $0.025 \times 3/n(s)$\\
    batch sizes for KAPAO-S, M, and L & $N_b$ & $128, 72, 48$\\
    pose, keypoint obj. conf. thresholds & $\tau_{cp}$, $\tau_{ck}$ & $0.001$, $0.2$\\
    pose, keypoint obj. $\iou$ thresholds & $\tau_{bp}$, $\tau_{bk}$ & $0.65$, $0.25$\\
    maximum fusion distance (px) & $\tau_{fd}$ & $50$\\
    pose obj. conf. threshold for fusion & $\tau_{fc}$ & $0.3$\\
    \hline
\end{tabular}
\smallskip
\caption[KAPAO hyperparameters.]{The hyperparameters used in the KAPAO experiments. $n(s)$ is the number of output grids.}
\label{tab:hyp}
\end{table}

\subsection{Influence of input size on accuracy and speed}
\begin{figure}[t]
\centering
    \includegraphics[trim={0cm, 0cm, 0cm, 0cm}, clip, width=0.6\linewidth]{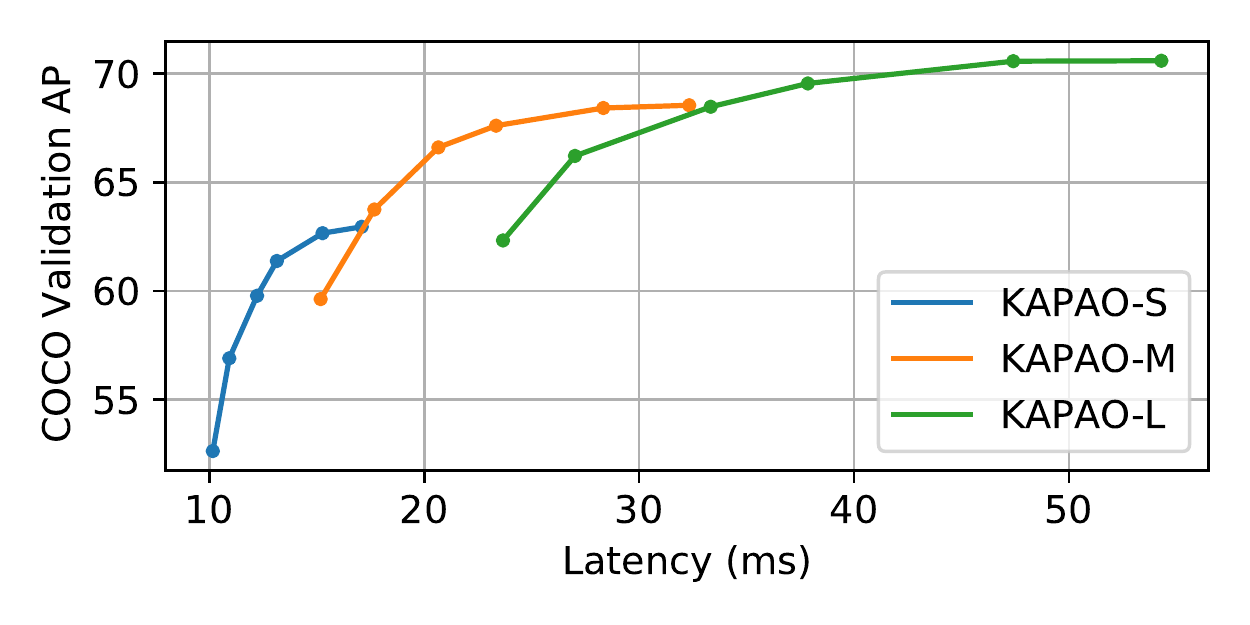}
\caption{Accuracy-speed trade-off for input sizes ranging from 640 to 1280. Evaluated on COCO \texttt{val2017} using a batch size of 1 on a TITAN Xp GPU.}
\label{fig:speed_size}
\end{figure}

The trade-off between accuracy and inference speed was investigated for various input sizes. The AP on COCO \texttt{val2017} was computed without TTA for $\max(w, h) \in \{640, 768, 896, 1024, 1152, 1280\}$. Figure~\ref{fig:speed_size} plots the results for each model. For all three models, reducing the input size to 1152 had a negligible effect on the accuracy but provided a meaningful latency reduction. For KAPAO-M and KAPAO-L, using an input size of 1024 reduced the accuracy marginally but also reduced the latency by $\sim$30\%.

\subsection{Video Inference Demos}
The source code includes five video inference demos. The first four demos run inference on RGB video clips sourced from YouTube to demonstrate the practical use of KAPAO under various inference settings. The final demo demonstrates the generalization of KAPAO by running inference on a depth video converted to RGB.  All reported inference speeds include \textit{all} processing (\ie, including image loading, resizing, inference, graphics plotting, \etc).

\begin{figure}
\centering
    \includegraphics[width=0.495\linewidth]{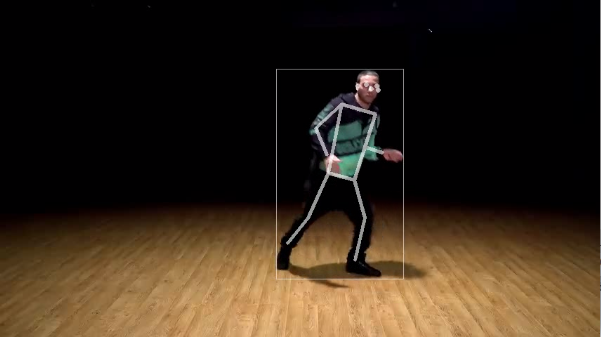}
    \includegraphics[width=0.495\linewidth]{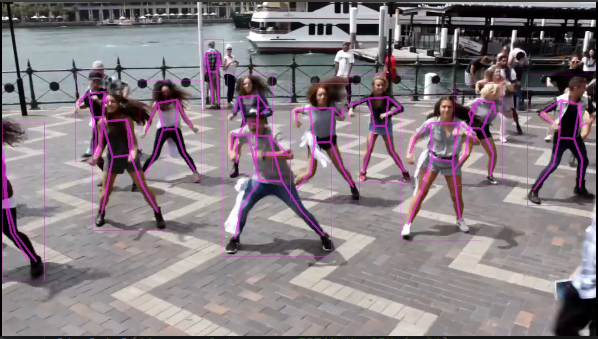}
    \includegraphics[width=0.495\linewidth]{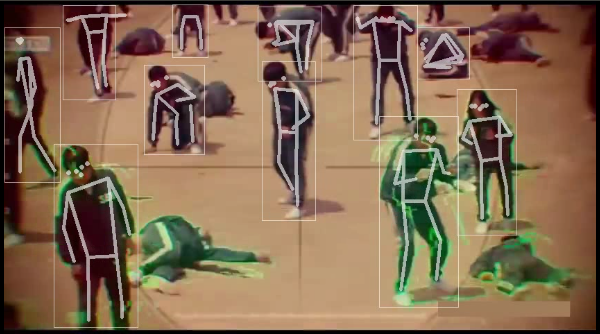}
    \includegraphics[width=0.495\linewidth]{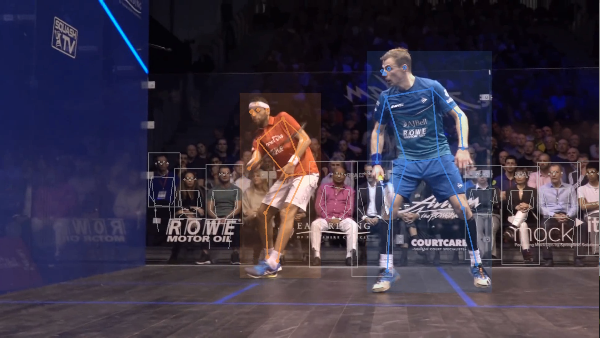}
\caption[KAPAO video inference screenshots.]{KAPAO video inference demo screenshots. Top-left: shuffling demo. Top-right: flash mob demo. Bottom-left: red light, green light demo. Bottom-right: squash demo. All video clips were sourced from YouTube.}
\label{fig:kapao_demos}
\end{figure}

\medskip\noindent\textbf{Shuffle.} KAPAO runs fastest on low resolution video with few people in the frame. This demo runs KAPAO-S on a single-person 480p dance video using an input size of 1024. The inference speed is $\sim$9.5 FPS on a workstation CPU (Intel Core i7-8700K), and $\sim$65 FPS on the TITAN Xp GPU. A screenshot of the inference is provided in Figure~\ref{fig:kapao_demos} (top-left).

\medskip\noindent\textbf{Flash Mob.} KAPAO-S was run on a 720p flash mob dance video using an input size of 1280. A screenshot of the inference is shown in Figure~\ref{fig:kapao_demos} (top-right). On a workstation housing a TITAN Xp GPU, the inference speed was $\sim$35 FPS.

\medskip\noindent\textbf{Red Light, Green Light.} This demo runs KAPAO-L on a 480p video clip from the TV show \textit{Squid Game} using an input size of 1024. For this demo, incomplete poses were plotted using keypoint objects only by setting $\tau_{ck}$ = 0.01 (see Section~\ref{sec:kapao_limitations} for details). A screenshot of the inference is provided in Figure~\ref{fig:kapao_demos} (bottom-left). The GPU inference speed varied between 15 and 30 FPS depending on the number of people in the frame. 

\medskip\noindent\textbf{Squash.} KAPAO-S was run on a 1080p slow-motion squash video using an input size of 1280. A simple player tracking algorithm was implemented based on the frame-to-frame pose differences. The inference speed was $\sim$22 FPS on the TITAN Xp GPU. A screenshot is provided in Figure~\ref{fig:kapao_demos} (bottom-right).

\begin{figure}
\centering
    \includegraphics[width=0.45\linewidth]{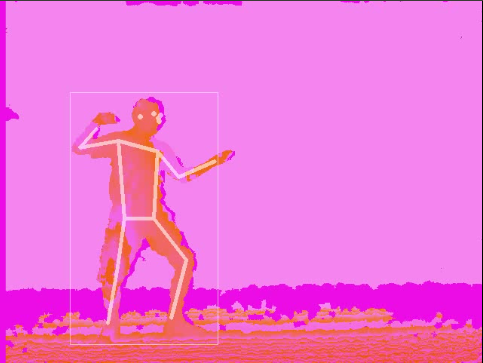}
\caption[KAPAO inference on a depth image.]{Pose objects generalize well and can even be detected in depth video. Shown here is a screenshot of KAPAO-S running inference on a depth video obtained from a fencing action recognition dataset~\cite{app_malawski2016classification, app_zhu2022fencenet}.}
\label{fig:kapao_depth}
\end{figure}

\medskip\noindent\textbf{Depth Videos.} Finally, the robustness and generalization capabilities of KAPAO are demonstrated by running inference with KAPAO-S on depth videos obtained from a fencing action recognition dataset~\cite{app_malawski2016classification}. The depth information was converted to RGB format in a 480p video. A screenshot from the inference video, which ran at $\sim$60 FPS on the TITAN Xp GPU, is displayed in Figure~\ref{fig:kapao_depth}. Despite the marked difference in appearance between the depth images and the KAPAO training images, human poses were still detected with high confidence. This interesting test result can be attributed to pose object representation learning, where spatial relations between human keypoints are learned using large-scale features and high-level context (\eg, like the edges making up the human shape). This is further supported by the fact that fewer keypoint objects were detected in the depth images. Zhu~\etal~\cite{app_zhu2022fencenet} use KAPAO to extract 2D pose information from depth videos and predict fine-grained footwork in fencing. 

\subsection{Error Analysis}
\label{sec:kapao_error}
While the AP and AR metrics are robust and perceptually meaningful (\ie, algorithms with higher AP/AR are generally more accurate), they can hide the underlying causes of error and are not sufficient for truly understanding an algorithm's behaviour. 
For example, the results presented in the previous section showed that KAPAO consistently provides higher AR than previous single-stage methods and higher AP at lower OKS thresholds (\eg, AP$^{.50}$).
The exact cause for this result cannot be understood through analysis of the AP/AR metrics alone, but a potential explanation is that KAPAO provides more precise person/pose detection (\ie, more true positives and fewer false positives and false negatives), but less precise keypoint localization.
To investigate this further, this section provides a more in-depth analysis of the error using the error taxonomy and analysis code provided by Ronchi and Perona~\cite{app_ronchi2017benchmarking}.

Ronchi and Perona propose an error taxonomy for multi-person human pose estimation on COCO that includes four error categories: \textit{Background False Positives}, \textit{False Negatives}, \textit{Scoring}, and \textit{Localization}. Scoring errors are due to sub-optimal confidence score assignment; they occur when two detections are in the proximity of a ground-truth annotation and the one with the highest confidence has the lowest OKS. Localization errors are due to poor keypoint localization within a detected instance; they are further categorized into four types: \textit{Jitter}: small localization error ($0.5 \leq \exp(-d_i^2/2s^2k_i^2) < 0.85$); \textit{Miss}: large localization error ($\exp(-d_i^2/2s^2k_i^2) < 0.5$); \textit{Inversion}: left-right keypoint flipping within an instance; and \textit{Swap}: keypoint swapping between instances. The reader may refer to~\cite{app_ronchi2017benchmarking} for a more detailed description of the error classifications. 

\begin{figure}[t!]
\centering
    \includegraphics[trim={0cm, -0.6cm, 0cm, 0cm}, clip, width=0.495\linewidth]{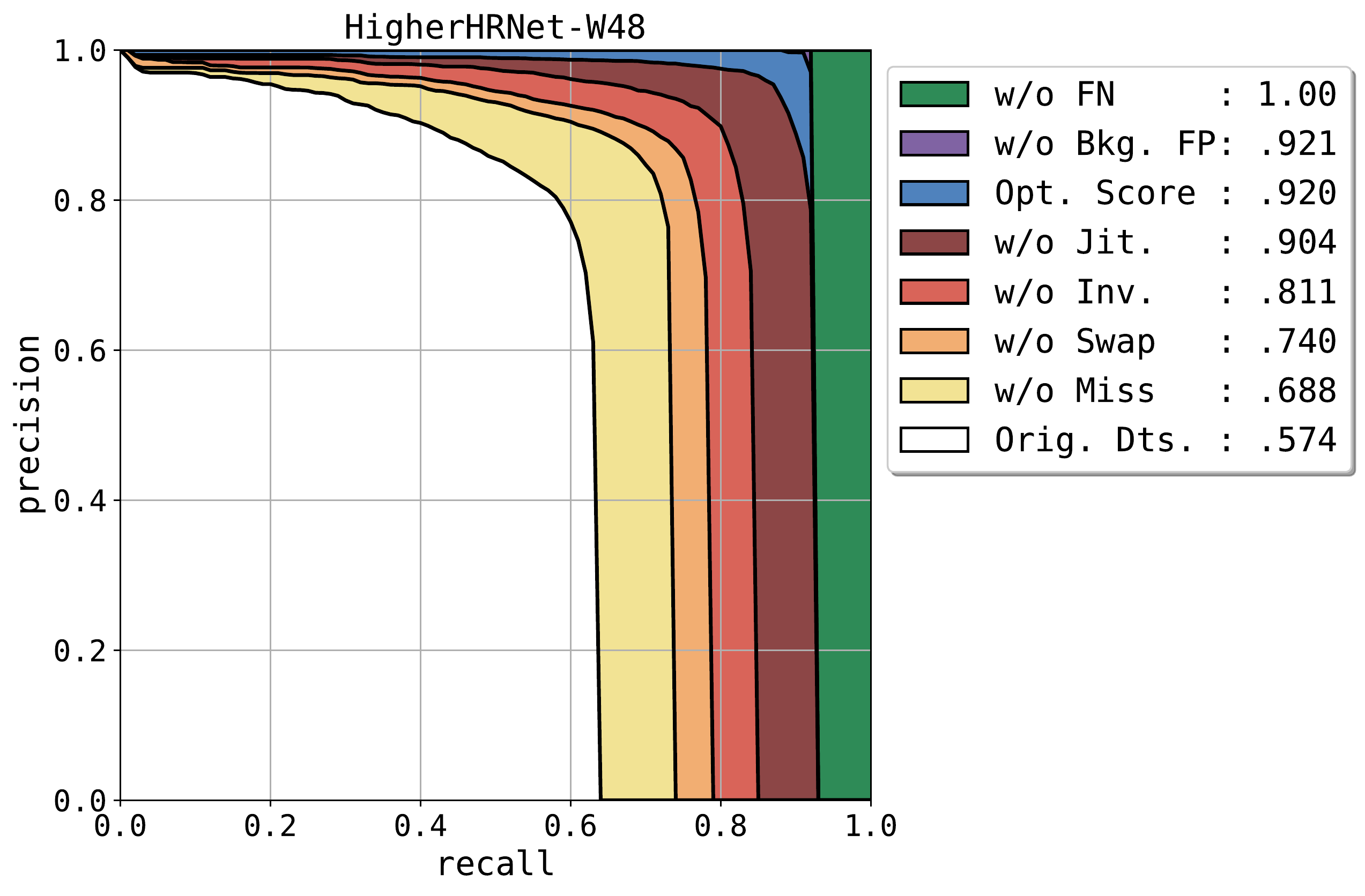}
    \includegraphics[trim={0cm, -0.6cm, 0cm, 0cm}, clip, width=0.495\linewidth]{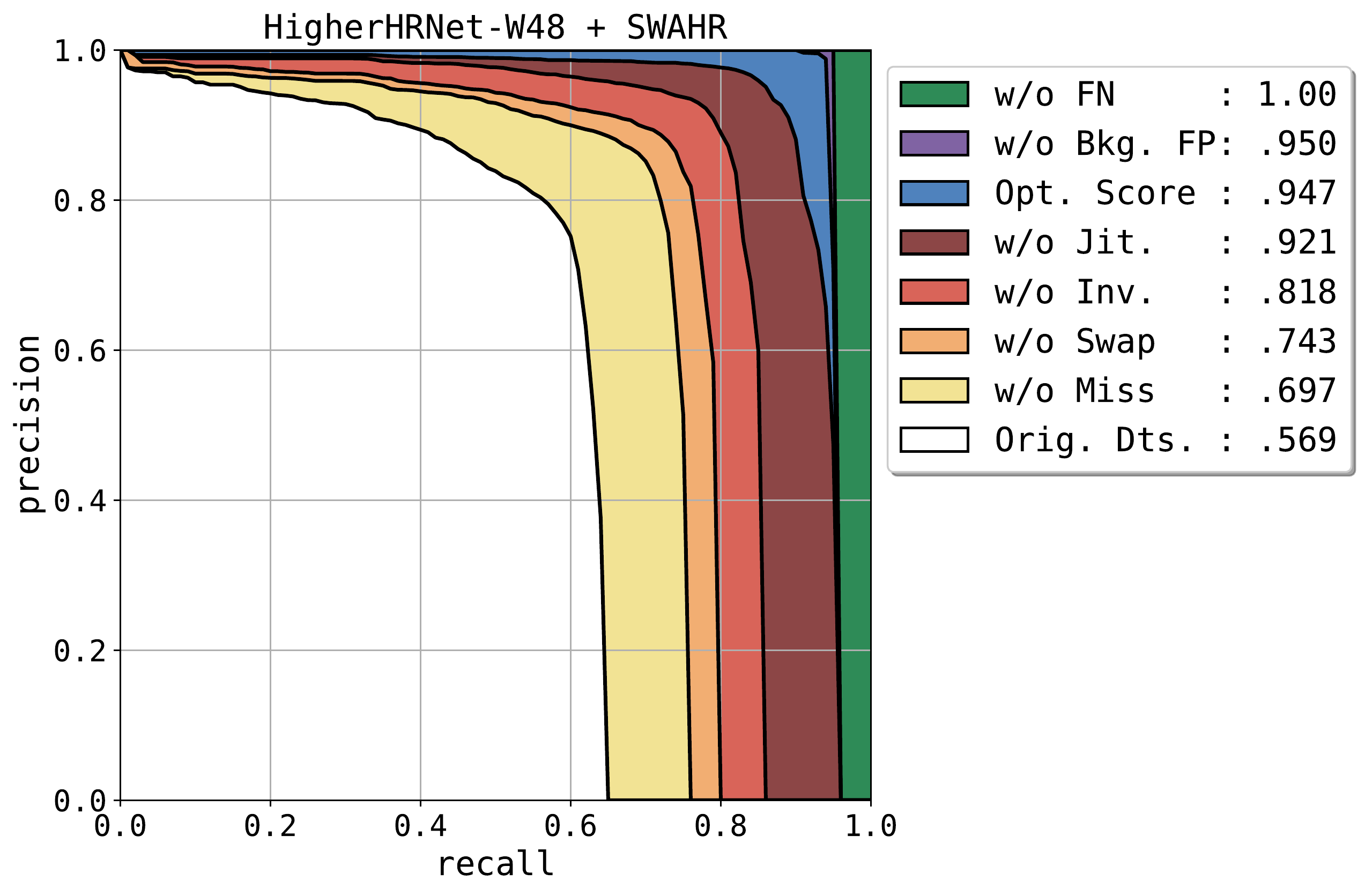}
    \includegraphics[trim={0cm, -0.6cm, 0cm, 0cm}, clip, width=0.495\linewidth]{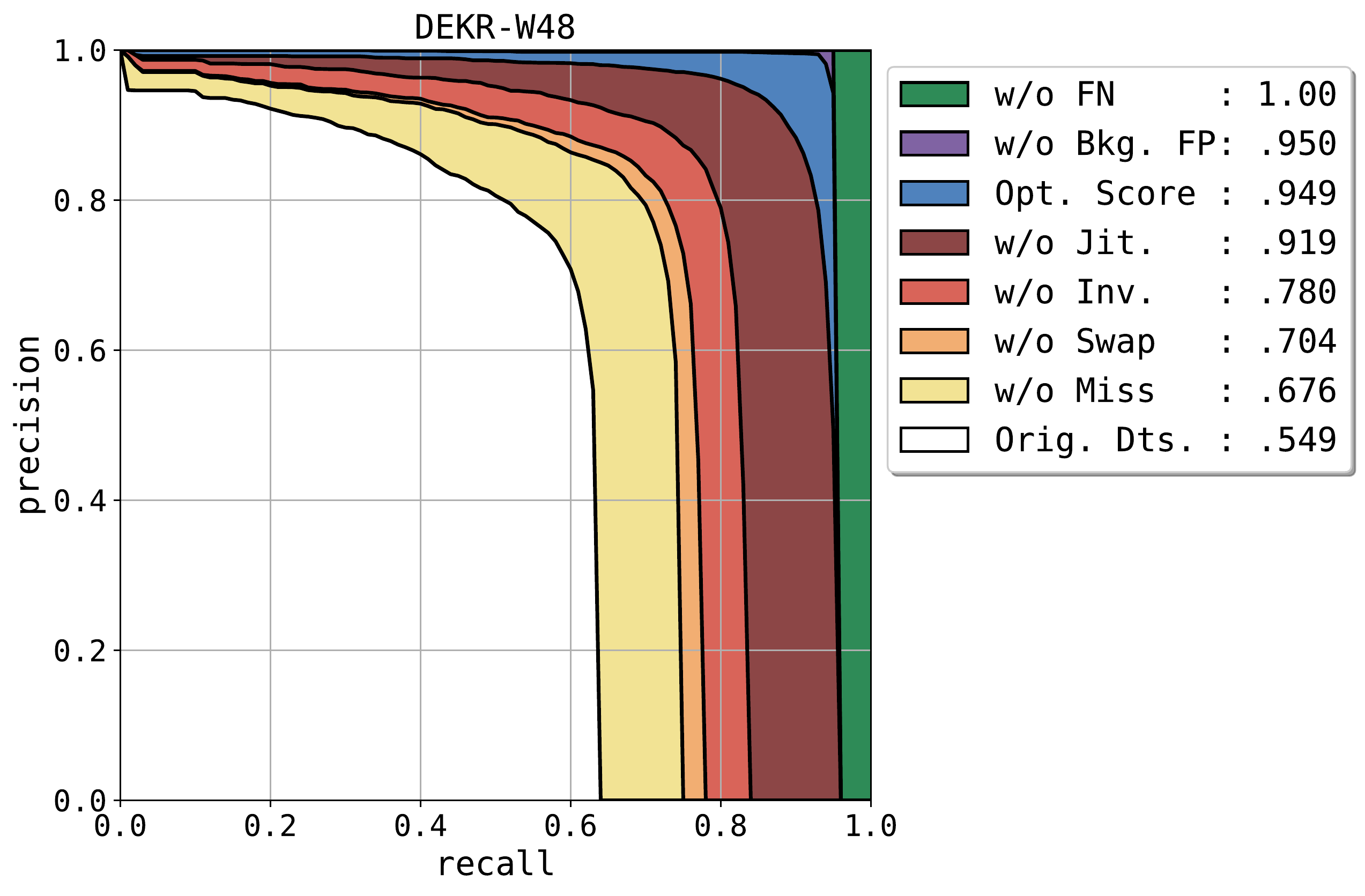}
    \includegraphics[trim={0cm, -0.6cm, 0cm, 0cm}, clip, width=0.495\linewidth]{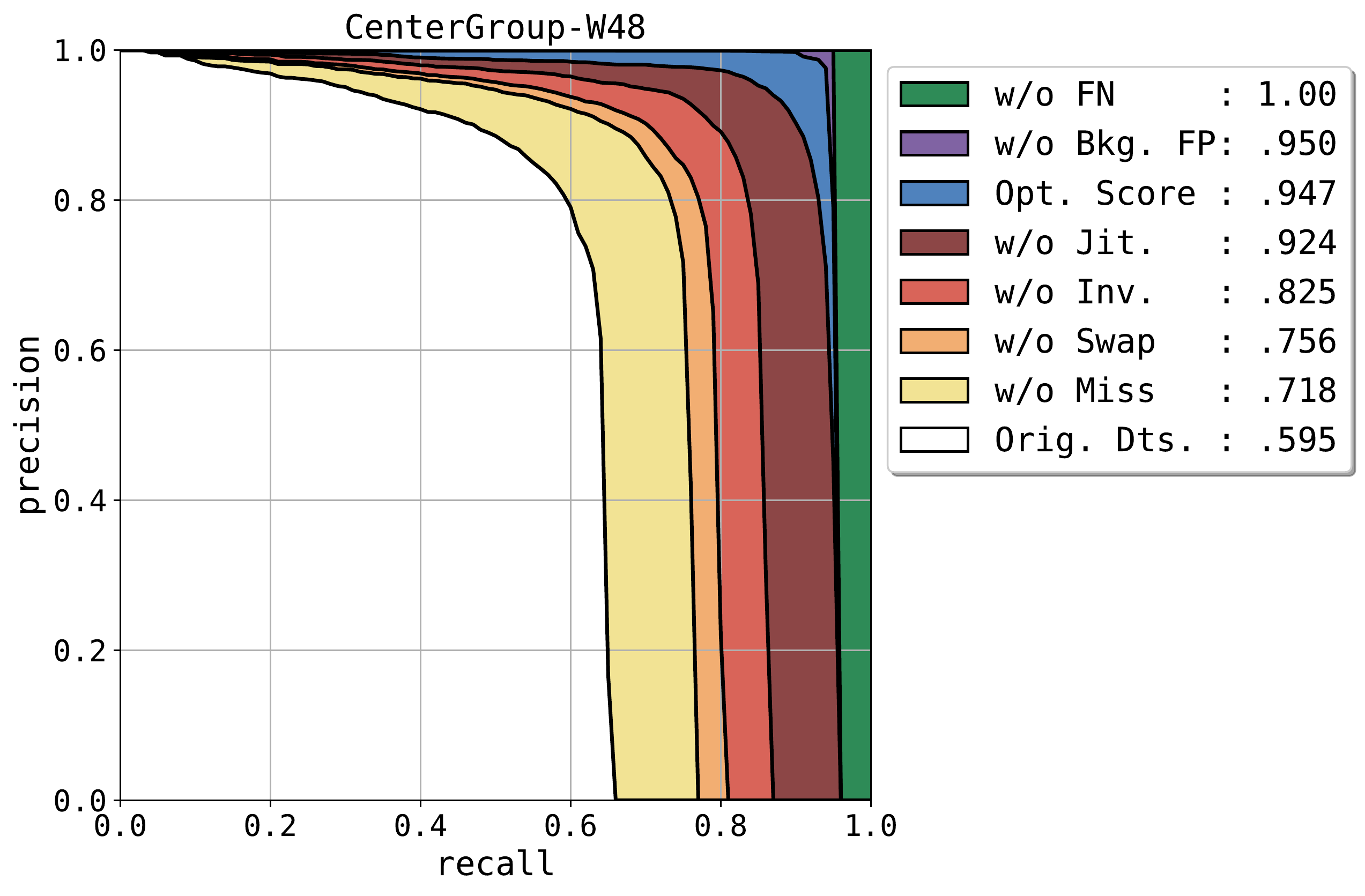}
    \includegraphics[trim={0cm, 0cm, 0cm, 0cm}, clip, width=0.495\linewidth]{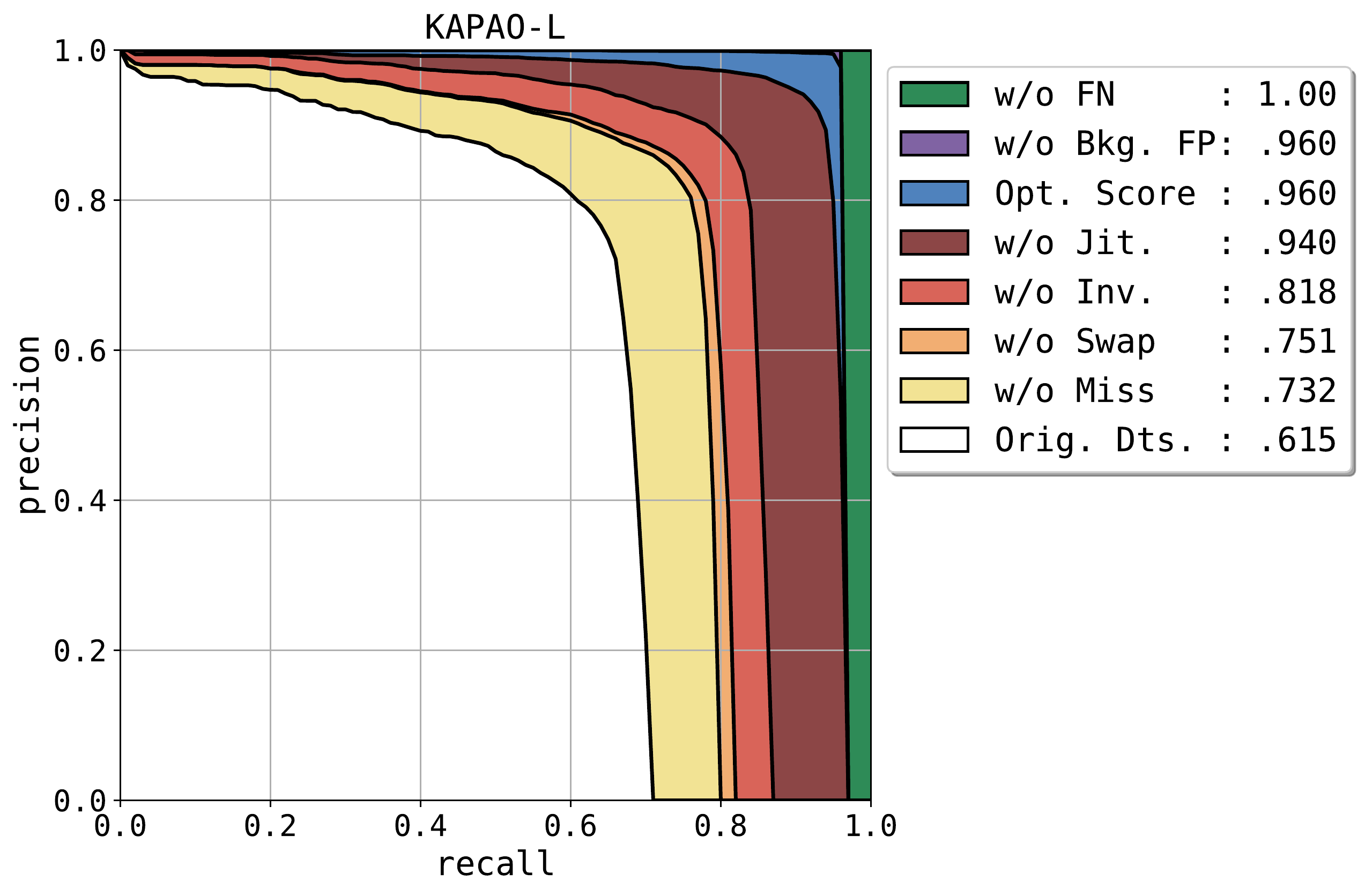}
\caption[Error type analysis on COCO \texttt{val2017}]{Error type analysis on COCO \texttt{val2017} for HigherHRNet-W48~\cite{cheng2020higherhrnet}, HigherHRNet-W48 + SWAHR~\cite{luo2021rethinking}, DEKR-W48~\cite{geng2021bottom}, CenterGroup-W48~\cite{braso2021center}, and KAPAO-L for an OKS threshold of 0.85 (without TTA). Plots generated using the \texttt{coco-analyze} toolbox~\cite{app_ronchi2017benchmarking}. AP$^{.85}$ values in legends given in decimals as opposed to percent.}
\label{fig:kapao_errors}
\end{figure}

\sloppy{
Figure~\ref{fig:kapao_errors} plots the precision-recall curves for HigherHRNet-W48~\cite{cheng2020higherhrnet}, HigherHRNet-W48 + SWAHR~\cite{luo2021rethinking}, DEKR-W48~\cite{geng2021bottom}, CenterGroup-W48~\cite{braso2021center}, and KAPAO-L at an OKS threshold of 0.85 using the results without TTA. Recalling that AP$^{\alpha}$ is equal to the area under the precision-recall curve generated at $\oks=\alpha$, the coloured regions in Figure~\ref{fig:kapao_errors} reflect the theoretical improvement in AP$^{.85}$ as a result of sequentially rectifying the aforementioned error types. 
}

KAPAO-L provides the highest original AP$^{.85}$ (represented by the white region). Furthermore, KAPAO-L is less prone to Swap and Inversion errors, which can be attributed to the pose object representation that models cohesive pose instances and eliminates the need for bottom-up keypoint grouping algorithms that are prone to such errors. This is further supported by DEKR having lower Swap error than the other three methods, which all perform bottom-up keypoint grouping. Interestingly, DEKR has the most room for improvement in assigning optimal confidence scores, which could be the motivation behind using a model-agnostic scoring regression network to improve the AP in the original paper (not included in these results). As previously hypothesized, KAPAO has more Jitter error than some of the other methods as a result of having less precise keypoint localization. Conversely, KAPAO-L provides better detection as shown by less improvement after correcting the false positive and false negative errors. It is speculated that since KAPAO was designed using an object detection network as its backbone, it is more optimized for person/pose object detection and less optimized for keypoint localization compared to the other single-stage human pose estimation methods. Rebalancing KAPAO to focus more on keypoint localization is thus a recommended area for future work.

\subsection{Qualitative Comparisons}
\label{sec:kapao_qualitative}
KAPAO-L predictions from COCO \texttt{val2017} are qualitatively compared with CenterGroup-W48 without TTA. To systematically review cases where KAPAO-L performs better than CenterGroup-W48, and \textit{vice versa}, the maximum OKS was found for each ground-truth instance ($\oks_{max}$) using the top-20 scoring pose detections for each model. For each validation image, the difference between the summations of the OKS maxima was computed:
\begin{equation}
\Delta\mathrm{OKS} = \Sigma\mathrm{OKS}_{max}^{KAPAO-L} - \Sigma\mathrm{OKS}_{max}^{CenterGroup-W48}
\end{equation}
$\Delta\oks$ is positive for images where KAPAO-L performs better than CenterGroup-W48. Conversely, $\Delta\oks$ is negative for images where CenterGroup-W48 performs better than KAPAO-L. To plot partial poses using KAPAO-L, the keypoint object confidence threshold $\tau_{ck}$ was lowered to 0.01 to promote the fusion of keypoint objects and increase the frequency of keypoint confidences in the predicted poses $\hat{\mathbf{P}}$ (see Section~\ref{sec:kapao_limitations} for details). The $\tau_{ck}$ of 0.01 lowered the AP from 70.4 to 70.1. It also increased the post-processing time from approximately 3 ms to 5 ms per image due to the increased number of keypoint objects that are fused in Algorithm~\ref{alg:fusion}. Using these inference settings, KAPAO-L is still 1.0 AP more accurate and 3.0$\times$ faster than CenterGroup-W48 on \texttt{val2017}. 

\begin{figure}[t!]
\centering
    \includegraphics[trim={0, 1cm, 0, 0}, clip, width=0.495\linewidth]{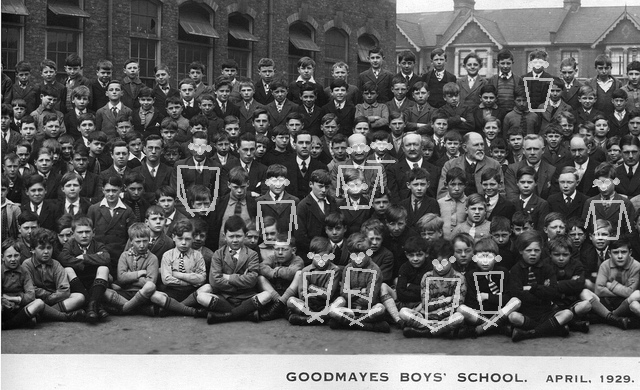}
    \includegraphics[trim={0, 1cm, 0, 0}, clip, width=0.495\linewidth]{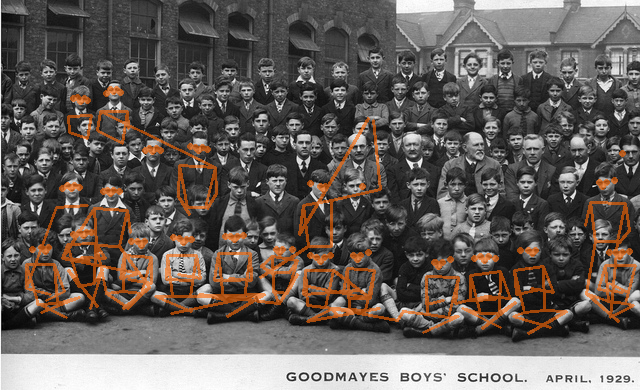}
    \includegraphics[trim={0, 1cm, 0, 0}, clip, width=0.495\linewidth]{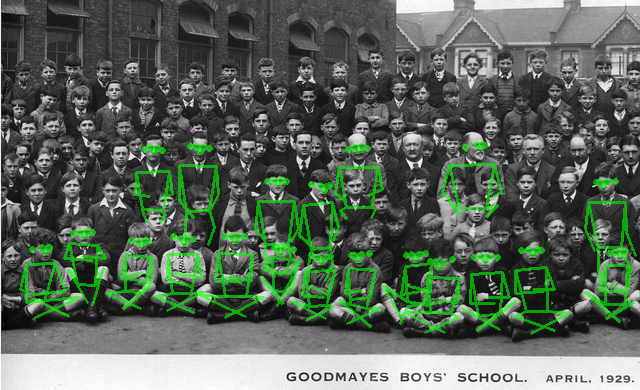}
    \includegraphics[trim={0, 1cm, 0, 0}, clip, width=0.495\linewidth]{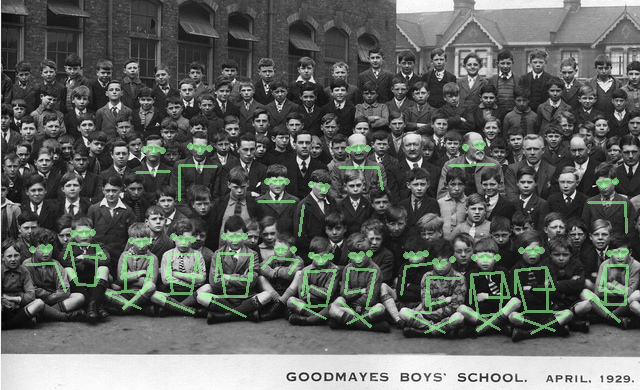}
\caption[Qualitative comparison between KAPAO-L and CenterGroup-W48 (COCO image \texttt{24021}).]{Qualitative comparison between KAPAO-L and CenterGroup-W48 (COCO image \texttt{24021}, $\Delta\oks=2.06$). Top-left: ground-truth. Top-right: top-20 scoring CenterGroup predictions. Bottom-left: top-20 scoring KAPAO predictions (all keypoints). Bottom-right: top-20 scoring KAPAO-L predictions (fused keypoint objects).}
\label{fig:kapao_qualitative_crowd}
\end{figure}

It is observed that extreme values of $\Delta\oks$ are associated with crowded images ($>$ 20 people) containing a limited number of annotations ($<$ 20 annotations). For these images, $\Sigma\mathrm{OKS}_{max}$ is contingent on whether the ground-truth instances are predicted by the top-20 scoring detections and therefore an element of chance is involved. 
Figure~\ref{fig:kapao_qualitative_crowd} illustrates such a scenario. The top-left image shows the ground-truth pose annotations (white); the top-right image shows the top-20 scoring CenterGroup-W48 detections (orange); the bottom-left image shows the top-20 scoring KAPAO-L detections (green); and the bottom-right image shows the same KAPAO-L detections but only plots the fused keypoint objects (light green). The top-20 KAPAO-L detections contain 8 of the 10 ground-truth instances whereas the top-20 CenterGroup-W48 predictions contain 6. As a result, $\Delta\oks = 2.06$. Because all the COCO keypoint metrics are computed using the 20 top-scoring detections, false negatives are artificially inflated while true positives are artificially deflated. While it is perplexing that the COCO metrics possess an element of randomness, it is conceivable that over many images the randomness averages out and does not favour one model over another. Moreover, the COCO dataset is sparsely populated with crowded images so these rare cases likely have a negligible influence on AP/AR. The implications of only using 20 detections on datasets like CrowdPose may be more severe and worth investigating, however.

\begin{figure}[t!]
\centering
    \includegraphics[trim={0, 0, 0, 4cm}, clip, width=0.495\linewidth]{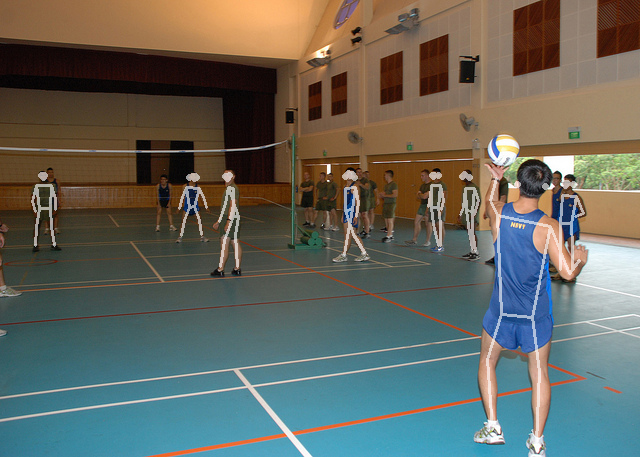}
    \includegraphics[trim={0, 0, 0, 4cm}, clip, width=0.495\linewidth]{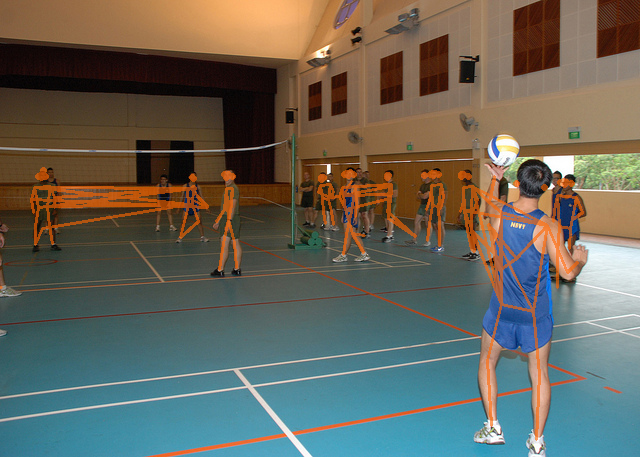}
    \includegraphics[trim={0, 0, 0, 4cm}, clip, width=0.495\linewidth]{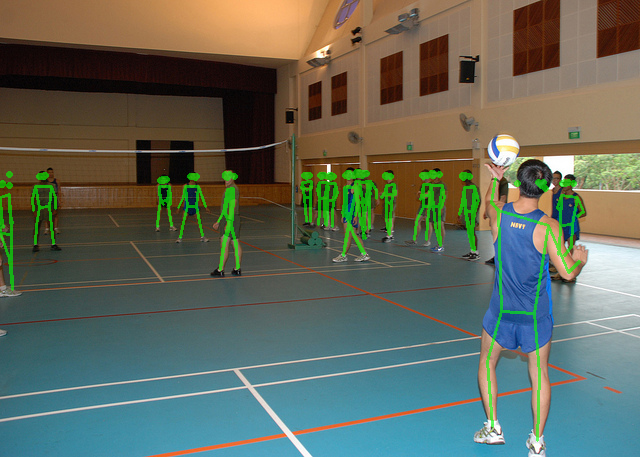}
    \includegraphics[trim={0, 0, 0, 4cm}, clip, width=0.495\linewidth]{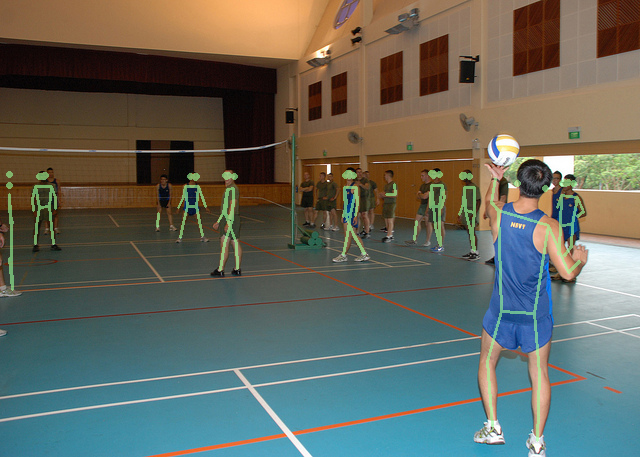}
\caption[Qualitative comparison between KAPAO-L and CenterGroup-W48 (COCO image \texttt{49759}).]{Qualitative comparison between KAPAO-L and CenterGroup-W48 (COCO image \texttt{49759}, $\Delta\oks=+0.68$). Top-left: ground-truth. Top-right: top-20 scoring CenterGroup predictions. Bottom-left: top-20 scoring KAPAO predictions (all keypoints). Bottom-right: top-20 scoring KAPAO-L predictions (fused keypoint objects).}
\label{fig:kapao_qualitative_pos}
\end{figure}

\begin{figure}[t!]
\centering
    \includegraphics[trim={0, 0, 0, 0}, clip, width=0.4965\linewidth]{figs/kapao_cg/0_000000326248_gt.png}
    \includegraphics[trim={0, 0, 0, 0}, clip, width=0.495\linewidth]{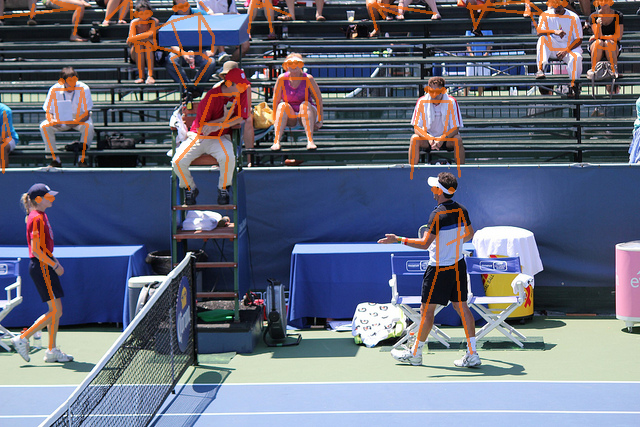}
    \includegraphics[trim={0, 0, 0, 0}, clip, width=0.495\linewidth]{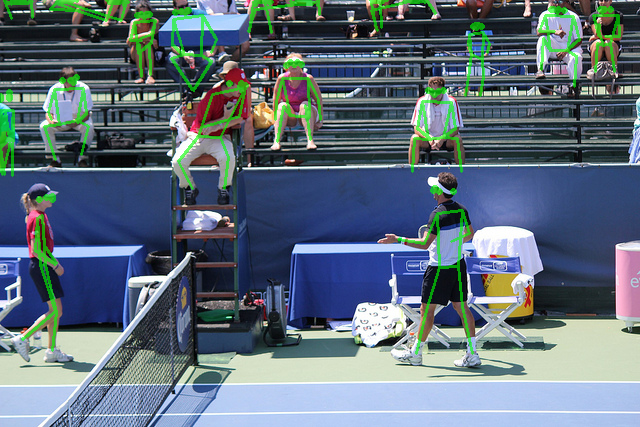}
    \includegraphics[trim={0, 0, 0, 0}, clip, width=0.495\linewidth]{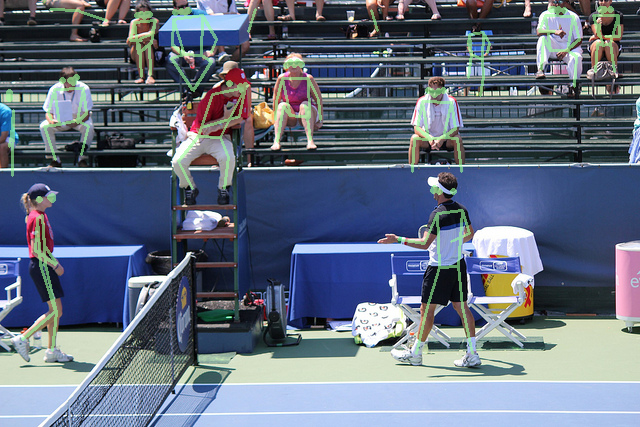}
\caption[Qualitative comparison between KAPAO-L and CenterGroup-W48 (COCO image \texttt{326248}).]{Qualitative comparison between KAPAO-L and CenterGroup-W48 (COCO image \texttt{326248}, $\Delta\oks=-0.66$). Top-left: ground-truth. Top-right: top-20 scoring CenterGroup predictions. Bottom-left: top-20 scoring KAPAO predictions (all keypoints). Bottom-right: top-20 scoring KAPAO-L predictions (fused keypoint objects).}
\label{fig:kapao_qualitative_neg}
\end{figure}

To avoid the aforementioned issues with comparing OKS in crowded scenes, the following comparisons consist of images where $\oks_{max}>0.5$ for all ground-truth instances using both models. Figure~\ref{fig:kapao_qualitative_pos} shows an example where KAPAO-L performs better than CenterGroup-W48 ($\Delta\oks=+0.68$). The keypoint grouping module of CenterGroup-W48 severely mixes up the keypoint identities (\textit{swap} error). Swap error is a common failure case for CenterGroup but an uncommon failure case for KAPAO due to its detection of holistic pose objects (quantitative errors provided in Figure~\ref{fig:kapao_errors}). Figure~\ref{fig:kapao_qualitative_neg} shows the image with the lowest $\Delta\oks$ ($-0.66$). For three of the ground-truth instances situated near the top of the frame, KAPAO predicts the locations of the nose, eyes, and ears significantly lower than the ground-truth locations, resulting in lower OKS for these poses. These errors are the result of the keypoint object bounding boxes being cut-off by the edge of the frame such that the center of the keypoint object bounding box no longer coincides with the actual keypoint locations. These errors could be rectified with relatively simple alterations to the inference code in future work. 

\end{document}